\documentclass[10pt,journal,compsoc]{IEEEtran}

\usepackage[linesnumbered,ruled,vlined]{algorithm2e}
\usepackage{amsmath}
\usepackage{mathtools,amssymb,lipsum}

\usepackage{cuted}
\usepackage{dsfont}
\usepackage{amsmath}
\usepackage{msc}
\usepackage{tikz}
\usetikzlibrary{shapes,arrows}
\usepackage{tcolorbox}
\usepackage{ amssymb }
\usepackage{balance}
\usepackage{graphicx}
\usepackage[normalem]{ulem}
\usepackage{url,cite}
\usepackage{balance}
\usepackage{graphicx}
\usepackage{enumerate}
\usepackage{stmaryrd}
\usepackage{color}
\usepackage{multirow}
\usepackage{array}
\usepackage{float}
\usepackage[english]{babel}
\usepackage[utf8]{inputenc}
\usepackage[noend]{algpseudocode}
\newcolumntype{L}[1]{>{\raggedright\let\newline\\\arraybackslash\hspace{0pt}}m{#1}}
\newcolumntype{C}[1]{>{\centering\let\newline\\\arraybackslash\hspace{0pt}}m{#1}}
\newcolumntype{R}[1]{>{\raggedleft\let\newline\\\arraybackslash\hspace{0pt}}m{#1}}

\SetKwInput{KwInput}{Input}                
\SetKwInput{KwOutput}{Output}              

\newcommand\like[1]{\begin{picture}(1,1)
\ifnum0=#1\put(.5,.35){\circle{1}}\else
\ifnum10=#1\put(.5,.35){\circle*{1}}\else
\put(.5,.35){\circle{1}}\put(.5,.35){\circle*{.#1}}
\fi\fi\end{picture}}
 
\usepackage{amsmath}

\IEEEtitleabstractindextext{%
\begin{abstract}
The federated learning (FL) technique was  developed to mitigate data privacy issues in the traditional machine learning paradigm. While FL ensures that a user's data always remain with the user, the gradients are shared with the centralized server to build the global model. This results in privacy leakage, where the server can infer private information from the shared gradients. To mitigate this flaw, the next-generation FL architectures proposed encryption and anonymization techniques to protect the model updates from the server. However, this approach creates other challenges, such as malicious users sharing false gradients. Since the gradients are encrypted, the server is unable to identify rogue users. To mitigate these attacks, this paper proposes a novel FL algorithm based on a fully homomorphic encryption (FHE) scheme. We develop a distributed multi-key additive homomorphic encryption scheme that supports model aggregation in FL. We also develop a novel aggregation scheme within the encrypted domain, utilizing users' non-poisoning rates, to effectively address data poisoning attacks while ensuring privacy is preserved by the proposed encryption scheme. Rigorous security, privacy, convergence, and experimental analyses have been provided to show that FheFL is novel, secure, and private, and achieves comparable accuracy at reasonable computational cost. 
\end{abstract}

}

\begin{document}

\title{FheFL: Fully Homomorphic Encryption Friendly Privacy-Preserving Federated Learning with Byzantine Users}

\author{Yogachandran Rahulamathavan, Charuka Herath, \textit{Student Member, IEEE}, Xiaolan Liu, \textit{Member, IEEE}, Sangarapillai Lambotharan \textit{Senior Member, IEEE} and  Carsten Maple, \textit{Senior Member, IEEE}
\IEEEcompsocitemizethanks{ 
\IEEEcompsocthanksitem Y. Rahulamathavan, C. Herath,  X. Liu and S. Lambotharan are with the Institute for Digital Technologies, Loughborough University London, London, U.K. (e-mails: \{y.rahulamathavan, C.Herath, xiaolan.liu, s.lambotharan\}@lboro.ac.uk).
\IEEEcompsocthanksitem C. Maple is with Secure Cyber Systems Research Group, WMG, University of Warwick, CV4 7AL Coventry, United Kingdom. E-mail: CM@warwick.ac.uk.
}
}
\maketitle

\section{Introduction}\label{Section: Intro}
Federated learning (FL) allows distributed users to learn collaboratively from data, without explicitly exchanging it. This is essential for several reasons, including data privacy requirements and regulations that limit the use and exchange of data. In this collaborative learning scheme, each user holds a portion of the distributed training data, which cannot be accessed by the central server (also known as the aggregation server). The users send their local gradients to the server, which aggregates and updates the global model, and then sends it back to the users. However, it has been revealed that sensitive information can be inferred from local gradients \cite{FL_Privacy_Leak}, leading to the development of privacy-preserving federated learning techniques (PPFL).

The most notable of PPFL algorithms utilise secure aggregation \cite{Bonawitz2017,MicroFL_2022, Advanced_MicroFL_2023, PrivateAggIEEEIoT2022},  differential privacy \cite{DP1_IEEE_Industry_2019,DP2_IEEE_IoT_2020}, and fully homomorphic encryption (FHE) \cite{PPFL_CKKS_2022,PPFL_HE_2022}. Each of these techniques has its benefits and drawbacks which are discussed in Section \textbf{\ref{Section: RelatedWork}} in further detail. These techniques have been developed to ensure that individual user updates are masked, anonymized or encrypted before sending them to the server, preventing the server from inferring users' privacy-sensitive information from the local model updates. While these approaches solve the privacy issue related to information leakage, this inadvertently realizes a security threat, known as a poisoning attack. 

In a poisoning attack, malicious users sabotage collaborative learning by training the local model on fake or rogue data and sending their malicious model updates to the server.  It should be noted that poisoning attacks are difficult to detect even in non-PPFL settings \cite{Tolpegin2020,Matei2020,Krum2017,Yanyang2020}; hence, this is another active area of research in FL. Notable techniques to mitigate poisoning attacks include machine learning to detect outliers \cite{Tolpegin2020}, clustering using similarities \cite{ShieldFL_IEEE_TIFS_2022} and detecting anomalies \cite{IEEEJSecCommu2022}; using the performance analysis of global model against local model \cite{FLCert_IEEE_Inf_2022}; gradient clipping \cite{Yanyang2020}; and removing extreme values \cite{Krum2017}. It should be noted that these detection techniques require the server to access the users' model updates in plaintext, which leads to the initial privacy issue related to information leakage. Therefore, solving privacy issues leads to security issues and vice versa.

\begin{table*}[]
\centering
\caption{Comparision between our work and previous works focussing on security and privacy in FL. \like{10} denotes strong yes and \like{0} denotes strong no.}.
    \label{Table: Literature Comparision}
\begin{tabular}{|l|c|c|c|c|c|c|c|}
\hline
\multicolumn{1}{|c|}{Scheme} & \multicolumn{1}{c|}{Year} & \multicolumn{1}{c|}{\begin{tabular}[c]{@{}c@{}}Single \\ Server\end{tabular}} & \multicolumn{1}{c|}{\begin{tabular}[c]{@{}c@{}}Two \\ Servers\end{tabular}} & \multicolumn{1}{c|}{\begin{tabular}[c]{@{}c@{}}Server side\\ Complexity\end{tabular}} & \multicolumn{1}{c|}{\begin{tabular}[c]{@{}c@{}}User side\\ Complexity\end{tabular}} & \multicolumn{1}{c|}{\begin{tabular}[c]{@{}c@{}}Communicational\\ Overhead for Users\end{tabular}} &  \multicolumn{1}{c|}{\begin{tabular}[c]{@{}c@{}}Weak Security\\ Assumption\end{tabular}} \\ \hline
So et al.,  \cite{Byzentine_2020}       & 2020     &        \like{10}         &  \like{0}     &    \like{5}       &      \like{10}                   &    \like{10}      &  \like{0}   \\ \hline
Lie et al.,  \cite{PEFL_2021}    & 2021 &    \like{0}&  \like{10}& \like{5}   & \like{2}   & \like{1}  & \like{5} \\ \hline
Z. Ma el al.,  \cite{ShieldFL_IEEE_TIFS_2022}     & 2022&\like{0} &\like{10} &\like{10} &\like{5} &\like{1} &\like{0} \\ \hline
H. Ma et al.,  \cite{MUD-PQFed2022}& 2022& \like{10}& \like{0}&\like{5} &\like{2} &\like{10} &\like{5} \\ \hline
Jebreel et al.,  \cite{FFL2023}& 2023&\like{10} &\like{0} &\like{2} &\like{2} &\like{10} &\like{5} \\ \hline
Zhang et al.,  \cite{LSFL}& 2023&\like{0} &\like{10} &\like{2} &\like{2} &\like{2} &\like{10} \\ \hline
FheFL (This work)     & 2024      &   \like{10}     &    \like{0}      &   \like{10}    &     \like{2}       &    \like{2}   &    \like{0}      \\ \hline
\end{tabular}
\end{table*}
In this particular context, we aim to develop a robust mechanism called FheFL that prevents malicious actors' intention to corrupt the FL process and ensures that each user's data remains secure and private throughout the entire process. While numerous works have tried to solve one of the two attacks, only a handful of works in literature have attempted to solve both of these attacks using a single solution \cite{Byzentine_2020,PEFL_2021,ShieldFL_IEEE_TIFS_2022,MUD-PQFed2022, FFL2023, LSFL}. However, as detailed in \textbf{Section \ref{Section: RelatedWork}}, these solutions either rely on two non-colluding servers \cite{PEFL_2021,ShieldFL_IEEE_TIFS_2022,LSFL} or collaboration between users \cite{FFL2023}, or follow \textit{all-on-nothing approch} i.e., punish the user (or the group that the user belongs to) by ignoring the model updates if the distance between the updated values is high \cite{Byzentine_2020,MUD-PQFed2022}. In contrast to the existing works, FheFL proposes a novel PPFL mechanism that leverages FHE to achieve robustness, security and privacy preservation. Moreover, FheFL utilizes only one server for secure aggregation. We also introduce a recursive weighted update scheme, that works well with FHE, to detect and mitigate the impact of poisoning users \textit{without} interacting with other user groups, hence overcoming both the all-or-nothing limitations and the need for interaction between users.

\subsection{Notation}
We use bold lower-case letters such as $\mathbf{x}$, to denote column vectors; for row vectors, we use the transpose $\mathbf{x}^T$. $||\mathbf{x}||$ denotes the vector norm where    $||\mathbf{x}||=\sqrt{\mathbf{x}^T\mathbf{x}}$. We use bold upper-case letters, such as  $\mathbf{A}$, to denote matrices and identify a matrix with its ordered set of column vectors. The set of real numbers is denoted by $\mathbb{R}$ and a real number matrix with $d \times d$ elements is denoted by $\mathbb{R}^{d \times d}$. We use $\mathds{Z}_q$ to denote the ring of integers  modulo $q$ and $\mathds{Z}_q^{n\times m}$ to denote the set of $n\times m$ matrix with entries in $\mathds{Z}_q$. An integer polynomial ring with degree $N$ is denoted by $\mathbb{Z}_q[X]/(X^N+1)$, where the coefficients of the polynomial are always bounded by $\mathbb{Z}_q$. We use $[ . ]$ to denote the ciphertext i.e.,  plaintext $x$ is encrypted into ciphertext $[x]$ where $x \in \mathbb{Z}_q[X]/(X^N+1)$.

\begin{figure*}[h]
  \centering
  \includegraphics[trim={3cm 3cm 3cm 3cm},width=1\textwidth]{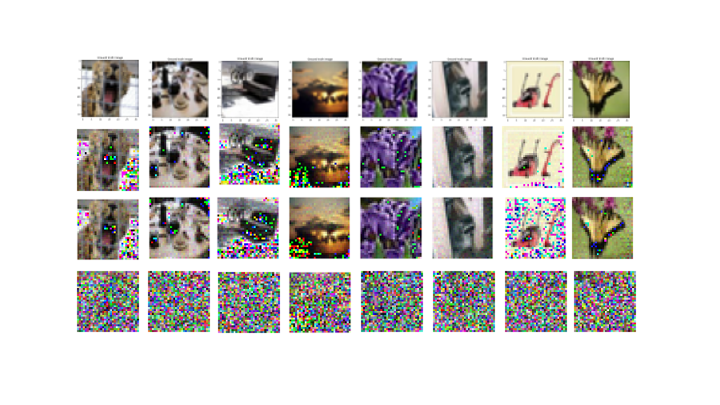}
  \caption{Reconstruction results of the Deep Leakage from Gradients (DLG) attack on CIFAR100 dataset. The first row shows the raw images from the CIFAR100 dataset. The second row displays reconstructed images when no gradient protection is applied. The third row shows the reconstructions when the gradients of the output layer are encrypted using the proposed scheme. The final row presents the reconstructed images when the gradients of the first hidden layer are protected with the proposed method.}
  \label{Figure: DLG}
\end{figure*}


\begin{tcolorbox}[colback=gray!10,colframe=gray!80!black]
\noindent \textbf{The key contributions are listed below:}

\textbf{1.} The paper proposes a non-poisoning rate-based weighted aggregation scheme that not only outperforms the other popular statistical approaches but integrates well with FHE schemes.

\textbf{2.} The paper introduces modifications to the existing CKKS-based FHE scheme, transforming it into a distributed multi-key additive homomorphic encryption (HE) scheme. This modification enables model aggregation \textit{without the need for two servers}.

\textbf{3.} The proposed FHE-based scheme eliminates privacy leakage by encrypting the gradients, preventing the server from inferring private information. Additionally, the non-poisoning rate-based aggregation scheme mitigates data poisoning attacks in the encrypted domain. The combined approach addresses both privacy and security concerns in FL.
\end{tcolorbox}

\section{Related Works}\label{Section: RelatedWork}

\subsection{Security Focused Approaches} 
FL is vulnerable to poisoning attacks by malicious users. To mitigate these attacks, statistical approaches such as Median  \cite{mean_median_2017}, Trimmed Mean \cite{mean_median_2017}, and KRUM  \cite{Krum2017} have been developed. The median approach calculates the median of all received model updates and then computes the mean of the updates that are closest to the median. The Trimmed Mean algorithm \cite{mean_median_2017} sorts the model updates based on Euclidean distance. A percentage of updates is then trimmed from both ends to remove outliers. The mean is calculated from the remaining updates, which serve as the global model. This approach balances robustness and inclusiveness in FL, mitigating poisoning data and enhancing model integrity. KRUM calculates the Euclidean distance between a user's gradients and all other updates, counting the number of times it is among the $k$ closest updates, and only aggregates the $k$ most trustworthy updates \cite{Krum2017}. Machine learning techniques, such as anomaly detection with principal component analysis (PCA), can also mitigate poisoning attacks \cite{Tolpegin2020}. 

All these approaches enforce the users to send their model updates in plain-domain to the server. Attempting to redesign these algorithms to process the data in the encrypted domain is computationally infeasible. Therefore, we proposed an FHE-friendly secure aggregation scheme in \textbf{Section \ref{S: Proposed Scheme}}. The experimental results show that the performance of the proposed algorithm is comparable to the popular schemes.

\subsection{Privacy Focused Approaches}
In FL, the server can extract sensitive information about users' private data from the shared gradients. Users can also orchestrate privacy attacks, but their effectiveness decreases with more participants. Server-side attacks are more potent, especially when the server can trace individual local updates back to their sources \cite{FL_Privacy_Leak}. Different types of privacy attacks exist, including membership inference, property inference, distribution estimation, and reconstruction attacks. To execute these attacks, the server needs access to individual local updates in plaindomain. 

There are several techniques are developed to thwart privacy attacks.  One common approach is using secure aggregation techniques, where the server can only access aggregated model updates, preventing the inference of individual user updates. Bonawitz et al. \cite{Bonawitz2017} proposed the first secure aggregation protocol for deep neural networks. In \cite{Bonawitz2017}, the users need to interact with all other users to generate a secret key between each user pair. Since this approach consumes significant communication bandwidth, a number of recent works \cite{MicroFL_2022, Advanced_MicroFL_2023} have been proposed to optimize the communicational overhead and to compensate for offline users.

One of the issues with the above secret sharing-based schemes is the communication complexity required to execute the secret key-sharing protocol i.e., each user needs to communicate with all other users (or all the users in a group). To overcome this limitation,  P. Zhao et al. \cite{PrivateAggIEEEIoT2022} proposed an elegant solution based on the computational Diffie–Hellman (CDH) problem where each user needs to communicate with its two neighbours only.

J. Ma et al.  \cite{PPFL_CKKS_2022} proposed a multi-user FHE scheme based federated aggregation scheme. The focus of this work is to improve the multi-user FHE scheme suitable for FL settings. On the other hand, J. Park et al. \cite{PPFL_HE_2022} proposed a third-party-assisted FHE-based scheme for FL. In this work, a third party generates pair of a public and private key pairs and distributes the public key to all the users and servers. The users encrypt the model updates using the FHE scheme and public key before sharing it with the server. The server aggregates the model updates from all the users in the encrypted domain. Then it will liaise with the third party to decrypt aggregated model. While this approach doesn't require secret sharing as in the previous scheme, the scheme requires trust in a third party.

While some methods adopt differential privacy (DP) \cite{DP1_IEEE_Industry_2019, DP2_IEEE_IoT_2020} by injecting noise into local updates before sending them to the server, it should be noted that DP offers reliable privacy guarantees primarily for small values of $\epsilon$. A lower $\epsilon$ value corresponds to a stronger privacy assurance. However, due to the added noise, DP-based FL approaches significantly compromise the overall model's accuracy. Consequently, privacy is obtained at the expense of accuracy.

\subsection{Security and Privacy Focused Approaches}
There are a few notable works that intend to achieve both privacy and security in FL \cite{Byzentine_2020,PEFL_2021,ShieldFL_IEEE_TIFS_2022,MUD-PQFed2022, FFL2023, LSFL}. Table \ref{Table: Literature Comparision} compares the proposed work and other recent works. The first well-known approach that tackles both the security and privacy issues in FL using a single aggregation server was published in \cite{Byzentine_2020}. In \cite{Byzentine_2020}, each user will be interacting with all other users to compute the distances between their local model against other users' model updates using HE for every epoch. Then each user sends its local model updates and the distances to the server for aggregation. The secure-two-party computation between the users at every epoch in \cite{Byzentine_2020} adds significant communication overhead and computational complexity to each user.

To offload the communication overhead and computational complexity from the users to the server, the works in \cite{PEFL_2021, ShieldFL_IEEE_TIFS_2022, LSFL} utilise two non-colluding servers. In \cite{PEFL_2021}, one of the two servers is considered a trusted server that possesses the secret keys of the users.  This server can decrypt intermediate encrypted parameters; hence the work in \cite{PEFL_2021} follows a weak security assumption. Moreover, there is a secure two-party computation between the two servers using additively HE using Paillier cryptography.

To overcome the limitations in \cite{PEFL_2021}, Z. Ma et al. \cite{ShieldFL_IEEE_TIFS_2022} proposed a scheme called ShieldFL using a two-trapdoor Paillier HE scheme. The ShieldFL also utilises two servers, where the servers need to perform secure two-party computation to detect poisoning users. The Paillier HE-based secure two-party computation in \cite{ShieldFL_IEEE_TIFS_2022} encrypts the gradient vector element by element i.e., if there are $m$ elements in a gradient vector, then there will be $m$ Paillier encryption operations and $m$ ciphertexts. Therefore, for a given security level (e.g., 128-bit or 256-bit security), the Paillier encryption approach used in \cite{ShieldFL_IEEE_TIFS_2022} generates a large number of ciphertexts and consumes a high computational cost compared to single-instruction-multiple-data (SIMD) approaches \cite{Rahul_IoT}. For the proposed FheFL  in our paper, we use CKKS based FHE scheme which exploits the SIMD technique to pack as many plaintext elements in one polynomial hence producing a compact ciphertext with just one encryption.

Recently another work \cite{LSFL} proposed a lightweight FL scheme called LSFL for edge computing use case. The LSFL scheme uses two servers to perform a global model aggregation while preserving the privacy of user model updates via a lightweight randomisation technique. While the LSFL algorithm is the most efficient scheme compared to all other schemes, it relies on a weak security assumption that no users would collude with a server. This is a weak assumption  given that there would be a number of malicious users in FL and it is possible for one of the servers to launch a Sybil attack on the LSFL scheme.

 To mitigate the two-server architecture proposed in the above works, the works in \cite{MUD-PQFed2022,FFL2023} proposing PPFL schemes secure against byzantine attacks with just one server. The common theme across these two works (similar to the above work in \cite{Byzentine_2020}) is to rely on users who communicates with all other users via two-party computations to identify poisoning users. H. Ma et. al.   proposed a scheme that uses secret sharing to achieve privacy and commitment scheme, and group-based averaging to detect malicious users \cite{MUD-PQFed2022}. In \cite{MUD-PQFed2022}, the quantized neural network has been considered to ensure that the average of the aggregated parameter is within a certain range. If the average falls outside the range, then the update is excluded from the global model. The drawback of this method is that if one user in a group is malicious then the whole group will be excluded from the global model updates. Moreover, since the scheme's security relies on the users in a small group, the number of users required to collude with the server to compromise a genuine user's model update is less.

Recently, Jebreel et al.,  proposed a novel FL architecture called fragmented FL \cite{FFL2023}. In \cite{FFL2023}, a user  interacts with another user via a lightweight two-party secure computation scheme to exchange the fragments of the model updates to mix the model updates. Therefore, only around 50\% of the elements in the model update vector belongs to the user and the remaining 50\% are coming from another user. This approach breaks the link between the user's training data and the shared model update hence preventing the inference attack by the server. One of the drawbacks of this approach is the communication overhead among the users in every epoch. Moreover, since the server can obtain the aggregated model updates of two users, the number of users required to collude with the server to compromise a genuine user's model update is just one, hence this scheme's security is based on weak security assumption. In contrast to existing work, our proposed scheme is designed as a single-server model and eliminates the need for user interaction in each epoch.

\section{Background Information} \label{Section: Background}
In FL, the training data is distributed across many users and a server would facilitate the collaborative learning process between the users. Based on a problem, the server defines a suitable neural network architecture with a number of layers and neurons. Using this information, the server randomly initiates a global model $\mathbf{w}^0$ and distributes it to all users. At the $i$th epoch, the server sends the current global model $\mathbf{w}^i$ to the users and  each user independently calculates the loss using their local dataset as follows:
\begin{equation}\label{Eqn: Loss}
    L_u(\mathbf{w}^i) = \frac{1}{n_u} \sum_{j=1}^{n_u} l(\mathbf{w}^i, x_{u,j}),
\end{equation}
where $L_u$ denotes the user $u$'s local objective function, $l(.)$ denotes the loss function, $x_{u,j}$ denotes the user $u$'s $j$th data sample and $n_u$ denotes the total number of data samples at user $u$. Now user $u$ minimises the objective function $L_u(.)$ using the gradients of it's as follows:
\begin{equation}
    \mathbf{w}_u^i = \mathbf{w}^i - \eta^i.\nabla L_u(\mathbf{w}^i, \zeta^j_u),
\end{equation}
where $\eta^i$ is a learning rate at the $i$th epoch. Now the user $u$ shares the gradient $\nabla L_u(\mathbf{w}^i, \zeta^j_u)$ with the server. Once the server receives the updates from all the users, it gets an average of all the updates to obtain the global model $\mathbf{w}^{i+1}$ as follows:
\begin{equation}\label{Eqn: ModelUpdates}
    \mathbf{w}^{i+1} = \mathbf{w}^i - \eta^i \sum_{u=1}^U p^i_u.\nabla L_u(\mathbf{w}^i, \zeta^j_u),
\end{equation}
where $p^i_u$ is the weight assign to user $u$'s model updates during the $i$th epoch. It should be noted that $\sum_{u=1}^Up^i_u = 1$. We propose a novel method to obtain $p^i_u$ in Subsection \ref{Section: weigted agg}. This above process is repeated until the model convergence.
\subsection{Attack Models}\label{Subsection: attack models}
\textbf{Privacy attack model :}  In terms of privacy attacks, we consider a scenario where a semi-honest server participates honestly in the protocol but attempts to extract sensitive information from the private local data of the participants during training. Although participants can also orchestrate privacy attacks based on global models, their effectiveness diminishes as the number of participants increases. On the other hand, server-side attacks are more potent, particularly when the server has access to individual local updates and can trace them back to their sources. Different types of privacy attacks exist, including membership inference attacks that aim to identify if a specific example was part of the training data, property inference attacks that try to deduce specific properties about the training data, distribution estimation attacks that seek to obtain examples from the same data distribution as the participants' training data, and reconstruction attacks that ambitiously attempt to extract the original training data from a participant's local update. To execute any of these attacks, the server requires access to individual local updates. Therefore, our objective is to thwart privacy attacks by preventing the server from obtaining the participants' original updates.

\textbf{Security attack model :} In the context of security attacks, the server has no control over the behavior of participants, allowing a malicious participant to deviate from the prescribed training protocol and launch attacks against the global model. Security attacks can be categorized as untargeted attacks and targeted attacks, each with distinct objectives. Untargeted attacks, such as model poisoning, aim to disrupt model availability, preventing the model from converging effectively.  In contrast, targeted attacks such as data poisoning via label flipping attacks, focus on compromising model integrity, tricking the global model into making incorrect predictions for specific inputs chosen by the attacker. 

In our research, we assume that no more than $20\%$ of the participants in the system can be attackers. Although some studies consider higher percentages of attackers, it is highly improbable to encounter more than $20\%$ attackers in real-world FL scenarios \cite{FFL2023}. For instance, in the case of Gboard with millions of users, controlling even a small percentage of user devices would require compromising a substantial number of devices, demanding significant effort and resources, thus making it impractical. Since the targetted attacks (i.e., data poisoning attacks) are subtle and lead to a false sense of security \cite{Tolpegin2020}, we experiment with the proposed algorithm against targeted attacks involving label-flipping. Furthermore, we assume that the attacker(s) have no control over the server or the honest participants.

\subsection{CKKS Fully Homomorphic Encryption} \label{Subsection: FHE} 
FHE schemes support both addition and multiplication in the encrypted domain \cite{Gentry}.  Since Gentry's groundbreaking work \cite{Gentry}, there are numerous improvements were made by several researchers to improve efficiency and scalability. Currently, FHE has reached an inflexion point where several relatively complex algorithms can be evaluated in an encrypted domain in near-real time \cite{CryptoNet,DiNN}. Single-instruction-multiple-data (SIMD) is one of the powerful techniques that has enhanced the efficiency of FHE by more than three orders of magnitude \cite{SIMD}. While there is a handful of FHE schemes, this paper focuses on the lattice-based    Cheon-Kim-Kim-Song (CKKS) FHE scheme \cite{CKKS} since it is the most efficient method to perform approximate homomorphic computations over vectors with real and complex numbers. 

The foundation of lattice-based HE schemes, including CKKS, is the Learning With Error (LWE) problem \cite{RegevLWE2009}. The LWE problem is to distinguish noisy pairs $(\mathbf{a}, b)=(\mathbf{a}, \mathbf{a}^T \mathbf{s}+e)$ from randomly generated pairs $(\mathbf{r}_1 \in \mathbb{Z}^n_q, r_2 \in \mathbb{Z}_q)$, where $\mathbf{a}, \mathbf{s} \in \mathbb{Z}_q^n$, $\mathbf{a}$ is uniformly sampled, $\mathbf{s}$ is secret, and  $e \in \mathbb{Z}_q$ is a small noise.  $\mathbf{a}$ can be public. Without the addition of the small noise $e$, the problem would have been solved easily with the Gaussian elimination technique. Using LWE, a message $x \in \mathbb{Z}_q$ can be encrypted into $[x] =\mathbf{a}^T \mathbf{s}+e +x$ and $[x]$ can be decrypted as $x \approx [x] - \mathbf{a}^T \mathbf{s}$. Since LWE-based encryption schemes are not efficient, a Ring-LWE (RLWE) problem was introduced in \cite{RLWE}, which is a variant of LWE but on rings. Instead of using vectors from $\mathbb{Z}_q^n$, the RLWE problem works on polynomial rings $\mathbb{Z}_q[X]/(X^N+1)$. Therefore, similar to LWE, if $a, s, e \in \mathbb{Z}_q[X]/(X^N+f1)$ then it is infeasible to distinguish the noisy pair $(a\cdot s + e, a)$ from  $(r_1, r_2)$ where $r_1$ and $r_2$ are uniformly sampled from  $\mathbb{Z}_q[X]/(X^N+1)$ \cite{RLWE}. Let's see how this RLWE hard problem is used to develop modern FHE schemes. Denote a plaintext as $x$ and the corresponding ciphertext as $[x] = (c_0, c_1)$ where the ciphertext components $c_0$ and $c_1$ can be defined as
\begin{equation}\label{Eqn: FHE} 
c_0 = a \cdot s+x+e, \;\;\; c_1 = a,
\end{equation}
where $s$ is secret, $a$ is a random  and publicly known  and $e$ is a small random ring elements. It should be noted that $x, a, s, e, c_0$ and $c_1 \in \mathbb{Z}_q[X]/(X^N+1)$. The encrypted values can be decrypted as follows:
\begin{equation}\label{Eqn: FHE Dec}
 x \approx c_0 - c_1 \cdot s.
\end{equation}
Since $e$ is small, the decryption process outputs a correct result. To show that this scheme can support both addition and multiplication in the encrypted domain let us denote another plaintext as $x'$ and the corresponding ciphertext as $[x'] = (c_0', c_1')$ where $c_0' = a' \cdot s+x'+e'$ and $c_1' = a'$. If the ciphertexts are sent to a server, the server can correctly compute the addition (i.e., $[x+x']$) and multiplication (i.e.,$ [x \cdot x']$) correctly in the encrypted domain. It should be noted that the server does not have access to the secret key $s$. 

\subsubsection{Homomorphic Addition}\label{Subsection: HE Addition}
To correctly calculate the sum of the plaintexts in the encrypted domain, the server just adds the ciphertexts and gets $[x+x'] = (c_0+c_0', c_1+c_1')$. This can be decrypted to obtain the plaintext sum $x+x'$ as follows:
\begin{equation}
    (c_0 + c_0') - (c_1+c_1')s \approx x+x',
\end{equation}
where $e+e'$ is small and negligible. 

\subsubsection{Homomorphic Multiplication}\label{Subsection: HE Mult}
The multiplication of the two ciphertexts produces the following  ciphertext with three components: $(c_0 \cdot c_0', c_0 \cdot c_1'+c_0' \cdot c_1, c_1 \cdot c_1')$. This can be decrypted as follows to obtain the $x \cdot x'$:
\begin{equation}\label{Eqn: FHE mult}
     c_0 \cdot c_0'-(c_0 \cdot c_1'+c_0' \cdot c_1)s - c_1 \cdot c_1's^2  \approx x \cdot x'.
\end{equation}
Since $x \cdot e'+x' \cdot e+e \cdot e'$ is small and negligible the decryption process is correct. However, when multiplying two ciphertexts, the resultant ciphertext has three components, increasing the resultant ciphertext size. To reduce the components back to two,  FHE schemes use a technique called re-linearisation (denoted as $Relin$ below) to assist the server to compute the holomorphic multiplication without expanding the ciphertext size. To do this, the server needs a special key called an evaluation key from the user. The user generates the evaluation key $evk$ using the secret key as follows: $evk = (a'' \cdot s+e''+p \cdot s^2, a'')$ where $p$ is a large publicly known number. Now the re-linearisation process takes both multiplied ciphertext $(c_0 \cdot c_0', c_0 \cdot c_1'+c_0' \cdot c_1, c_1 \cdot c_1')$ and $evk = (a'' \cdot s+e''+p \cdot s^2, a'')$ to reduce the number of ciphertext components to two as follows:
\begin{eqnarray}
    \nonumber&&Relin((c_0 \cdot c_0', c_0 \cdot c_1'+c_0' \cdot c_1, c_1 \cdot c_1'), evk)\\
    \nonumber&=& ((c_0 \cdot c_0', c_0 \cdot c_1'+c_0' \cdot c_1) \\
    \nonumber &&- p^{-1} \cdot c_1 \cdot c_1'(a'' \cdot s+e''+p \cdot s^2, a''))=(c_0'', c_1''),
\end{eqnarray}
where $c_0''= c_0 \cdot c_0' - p^{-1} \cdot c_1 \cdot c_1' \cdot a'' \cdot s+p^{-1} \cdot c_1 \cdot c_1' \cdot e''+c_1 \cdot c_1' \cdot s^2$, and $c_1''=c_0 \cdot c_1'+c_0' \cdot c_1 -p^{-1} \cdot c_1 \cdot c_1' \cdot a''$. Now $x \cdot x'$ can be obtained by decrypting $(c_0'', c_1'')$ as $c_0'' - c_1''s$ which is equvalent to (\ref{Eqn: FHE mult}). This is correct since $p^{-1}c_1 \cdot c_1' \cdot e''$ is small and negligible.

CKKS FHE scheme is specially designed to support  real numbers and suitable FHE candidate for applications relying on vectors of real numbers. CKKS scheme introduced a novel encoding and decoding technique to convert  a message $\mathbf{x} \in \mathbb{R}^M$, a vector of real values into an integer polynomial $x \in \mathbb{Z}_q[X]/(X^N+1)$ where $M=N/2$ and vice versa. CKKS utilises a residue number system to speed up the computation. In the residue number system, a big integer is split into several small integers, and the addition and the multiplication of the original big integers are equivalent to the corresponding component-wise operations of the small integers i.e., $2^q = 2^{q_0 + q_1 + q_2 + \ldots + q_L+q_{L+1}}$ where $q_0 > q_i, i=1,\ldots L$, $q_{L+1} > q_i, i=1,\ldots L$, and $\Delta \approx q_1 \approx q_2 \approx q_3 \approx \ldots q_L$. In CKKS, this $L$ represents the number of multiplications that can be performed to a  ciphertext correctly.

 \section{FheFL: The Proposed Scheme}\label{S: Proposed Scheme}

We propose FheFL, a novel algorithm designed to address privacy and poisoning attacks in FL. To mitigate these threats, we introduce modifications to the CKKS FHE scheme, making it suitable for distributed networks utilising shared secret keys. Additionally, we develop a robust aggregation scheme based on users' non-poisoning rate, specifically designed to mitigate data poisoning in the encrypted domain. The fundamental concept behind FheFL is that each user employs a shared secret key to encrypt their model updates using the CKKS FHE scheme and transmits these encrypted updates to the server. By leveraging the homomorphic properties of CKKS, which enable computation on encrypted data, the server can aggregate the model updates from all users within the encrypted domain. Consequently, the resulting aggregated model remains encrypted under the secret key. To facilitate decryption of the aggregated model, each user sends a randomised shared key to the server to reconstruct the secret key from their respective shared secret keys. This cooperative effort enables the server to successfully decrypt the aggregated model. Below we first introduce the proposed  aggregation scheme followed by distributed secret sharing scheme, distributed CKKS FHE scheme and necessary encrypted domain functions.


\subsection{Non-poisoning rate based aggregation scheme} \label{Section: weigted agg}
The fundamental idea of this algorithm is to rely on aggregated global model update $\mathbf{w}^{i-1}$ from  $(i-1)$th epoch as a baseline to judge the model updates $\mathbf{w}^i_u$ from all the users in $i$th epoch. The server calculates a squared Euclidean distance $d_u^i$  between $\mathbf{w}^{i-1}(.)$ and $\mathbf{w}^i_u(.), \forall u=1,\ldots, U$ as follows:
\begin{eqnarray}
    \label{eqn: Euclidean Distance} d^i_u &=& ||\frac{\mathbf{w}^{i-1} - \mathbf{w}^i_u}{\eta^i}||_2^2 = ||\nabla L_u(\mathbf{w}^i, \zeta^j_u)||_2^2.
\end{eqnarray}
The Euclidean distance $d^i_u$ determines how far the model update from the user $u$ is changed compared to the global model obtained in the previous iteration. The assumption is that this distance will be very high if the model update is poisoned. However, this may not be always true if a user is trying to update the model using novel but legitimate data \cite{euc_dist_2023}. Therefore, instead of completely removing the users' updates when computing the new global model, we use the weight $d^i_u$ to minimise the impact of the poisoning attack. The intuition is to scale down the users' contribution to the global model in proportion to the user's Euclidean distance $d^i_u$. If $d^i_u$ is higher then the user $u$'s contribution to the global model should be smaller and vice versa.

To reflect the above intuition, let us define a new parameter called the user $u$'s \textit{non-poisoning rate}, denoted as $p^i_u$, during the $i$th iteration as follows:
\begin{equation}\label{Eqn: Poisoning rate}
    p^i_u = \frac{1}{U-1}.\left(1-\frac{d^i_u}{\sum_{j=1}^{U}d_i^j}\right),
\end{equation}
where $\frac{1}{U-1}$ is to ensure that the $\sum_{j=1}^{U} p_j^i = 1$. This can be confirmed by the following:
\begin{equation}\label{Eqn: Sum Poisoning rate}
  \sum_{u=1}^{U}(1-\frac{d^i_u}{\sum_{j=1}^{U}d_i^j}) = U-1.
\end{equation}
However, the server receives only the encrypted gradients $[\nabla L_u(\mathbf{w}^i, \zeta^j_u)]$  from all the users, therefore, we need to redesign (\ref{Eqn: ModelUpdates}), (\ref{eqn: Euclidean Distance}) and  (\ref{Eqn: Poisoning rate}) to support encrypted domain processing.  The server, who has access to the global model $\mathbf{w}^{i}$ in the plain domain, receives an encrypted input $[\nabla L_u(\mathbf{w}^i, \zeta^j_u)]$ from user $u$. Since each user uses its secret key to encrypt the model updates, we need to modify the CKKS FHE scheme suitable for distributed computations in the encrypted domain. Therefore, we introduce distributed key sharing and multi-key HE algorithms in the following subsections.

\subsection{Distributed Key Sharing}\label{Subsection: Distributed Key Sharing}
Secret sharing is a powerful technique that ensures a secret value can only be obtained when a minimum number of secret shares are combined. We adopt a pairwise secret-sharing scheme where a secret key, denoted as $s$, is shared among a group of users ($U$) \cite{SecretSHaring2011}. Each user ($u$) generates its own shared secret, $s_u$, such that $s=\sum_{u=1}^{U}s_u$. To enhance privacy in FL, user $u$ encrypts its locally trained model using its corresponding shared secret $s_u$ before transmitting it to the server. The server aggregates the model updates from all users in a manner that the aggregated model remains encrypted \textit{but} under the secret key $s$. Decrypting the aggregated model necessitates the server's possession of $s_u$ from each user, but since $s_u$ is utilized for encryption, it cannot be sent to the server directly. To address this challenge, each user engages in pairwise interactions with other users to generate a unique pairwise secret key between user $i$ and $j$. Let's denote the pairwise secret key between user $i$ and $j$ as $s_{i,j}$, where $s_{i,j} = -s_{j,i}$.

To assist the server in decryption, user $u$ transmits $ss_u = s_u+\sum_{j=1,\neq u}^{U}s_{u,j}$, where the user's shared secret key $s_u$ is masked by the pairwise secret keys contributed by other users. The server computes the secret key $s$ by summing all $ss_u$ as follows:
\begin{eqnarray}
   \nonumber \sum_{u=1}^{U}ss_u &=& \sum_{u=1}^{U}\{s_u+\sum_{j=1,\neq u}^{U}s_{u,j}\},\\
  \label{Eqn: mk shared secret} &=& \sum_{u=1}^{U}s_u+\underbrace{\sum_{u=1}^{U}\sum_{j=1,\neq u}^{U}s_{u,j}}_{= 0, \because s_{i,j}=-s_{j,i}}
   = \sum_{u=1}^{U}s_u = s. 
\end{eqnarray}
Note that $s_u$ can be randomly and freshly generated for each encryption. This computation ensures that the aggregated randomised shared secret keys match the original secret key, allowing for successful decryption of the aggregated model. It is important to note that the generation of pairwise secret keys between users only needs to be executed once when a user joins the network. Various protocols, such as the Diffie-Hellman key exchange, can be utilized to generate the pairwise secret keys, and the communication required for this process can be optimized as demonstrated in previous works like \cite{MUD-PQFed2022}, although this optimization is not the primary focus of this paper. Algorithm 1 provides the pseudo-code for the proposed distributed key-sharing protocols described above.

\begin{algorithm}[!ht]
\DontPrintSemicolon
  \KwInput{Security Level $n$}
  \KwOutput{Users Keys}
\tcc{System Initialisation}
  {Define $N$, $L$, $\Delta$, $q$ and $a$}\\
\tcc{User Setup}
  \For{User $u$}
    {
        Generate $a_u''$ and secret key  $s_u$\tcp*{}
        \For{User $j$}
        {
        User $u$ shares a secret key $s_{u,j}$ with user $j$ where $s_{u,j} = - s_{j,u}$ 
        }
    }
\caption{Setup and Key Distribution}
\end{algorithm}

\subsection{Multi-key homomorphic encryption scheme}\label{Subsection: Multikey FHE Scheme} 
Now we can put everything together to  build a multiuser HE scheme.  User $u$ encrypts a plaintext $x_u$ into a ciphertext $(c_0^u, c_1^u)$ where $c_0^u = as_u+x_u+e_u$ and $c_1^u=a$ where the public value $a$ is common for all the users. User $u$ also generates the evaluation key $evk_u = (a''_us_u+e''_u+ps_u^2, a''_u)$ to facilitate the relinearisation. Now user $u$ shares $(c_0^u, c_1^u, evk_u, ss_u)$ with the server. 
The server needs to obtain $\sum_{u=1}^{U}x_u$ from $(c_0^u, c_1^u, evk_u, ss_u)$ received from all the users. The following computation can achieve this:
\begin{eqnarray}
 \nonumber   \sum_{u=1}^{U}c_0^u   - \sum_{u=1}^{U}c_1^u.ss_u &=& \sum_{u=1}^{U}c_0^u   - a.\underbrace{\sum_{u=1}^{U}ss_u}_{=\sum_{u=1}^{U}s_u},\\
 \nonumber   &=& \sum_{u=1}^{U}c_0^u   - a.\sum_{u=1}^{U}s_u \approx \sum_{u=1}^{U}x_u.
\end{eqnarray}
This is correct since $\sum_{u=1}^{U}c_0^u =a\sum_{u=1}^{U}s_u+\sum_{u=1}^{U}x_u+\sum_{u=1}^{U}e_u$.

\subsection{Redesign (\ref{Eqn: ModelUpdates}), (\ref{eqn: Euclidean Distance}) and  (\ref{Eqn: Poisoning rate})  for encrypted domain processing}

The server received $[\nabla L_u(\mathbf{w}^i, \zeta^j_u)]$ which is encrypted under the user's secret key $s_u \in \mathbb{Z}_q[X]/(X^N+1)$ using the CKKS FHE scheme. Leveraging the properties of FHE, the server is able to perform computations on the encrypted data. Thus, the server can calculate $d^i_u$ in (\ref{eqn: Euclidean Distance}) in the encrypted domain as follows:
 \begin{eqnarray}
  \label{eqn: ED Euclidean Distance }[d^i_u] &=& [\nabla L_u(\mathbf{w}^i, \zeta^j_u)]^T.[\nabla L_u(\mathbf{w}^i, \zeta^j_u)].
\end{eqnarray}
This requires scalar multiplication in encrypted domain i.e., it requires the multiplication  of two  encrypted values followed by the addition of encrypted values. Both of these can be done easily using the description provided in Subsections \ref{Subsection: HE Addition} and \ref{Subsection: HE Mult}. It should be noted that the calculation of $[d^i_u]$ doesn't require input from other users therefore, $[d^i_u]$ is still encrypted under user $u$'s shared secret key.

However, to compute the non-poisoning rate $[p^i_u]$ for  user $u$ as per (\ref{Eqn: Poisoning rate}), the server needs to aggregate $[d^i_u]$s from all users which are encrypted under different keys. First of all, the server sums the squared Euclidean distances from all users and obtains $[\sum_{u=1}^{U}d^i_u]$. As described in Subsection (\ref{Subsection: Multikey FHE Scheme}),  $[\sum_{u=1}^{U}d^i_u]$ is now effectively encrypted by $\sum_{u=1}^{U}s_u$ i.e., the sum of the shared secret keys of all  users. The server can use the randomised shared secret keys from users to retrieve $\sum_{u=1}^{U}s_u$ hence it can decrypt and obtain $\sum_{u=1}^{U}d^i_u$. However, individual users squared Euclidean distances cannot be decrypted hence the non-poisoning rate of each can only be calculated in the encrypted domain i.e., as per (\ref{Eqn: Poisoning rate}), $[p^i_u ]$ is computed as follows:
\begin{equation}\label{Eqn: Poisoning rate in ED}
    [p^i_u] = \frac{1}{U-1}.\left(1-\frac{1}{\sum_{j=1}^{U}d_i^j}.[d^i_u]\right).
\end{equation}
In order to compute the global model as in (\ref{Eqn: ModelUpdates}), the server first computes $[p^i_u].[\nabla L_u(\mathbf{w}^i, \zeta^j_u]$ which is encrypted under user's shared secret key. Now using the multi-key HE described in Subsection \ref{Subsection: Multikey FHE Scheme}, the server can aggregate $[p^i_u].[\nabla L_u(\mathbf{w}^i, \zeta^j_u]$ for all users in the encrypted domain and decrypt its value using the shared secret keys. Then it multiplies it by $\eta^i$ to get $\eta^i \sum_{u=1}^U p^i_u.\nabla L_u(\mathbf{w}^i, \zeta^j_u)$. Now using this and the global model from previous iteration, the server can obtain $\mathbf{w}^{i+1}$ using (\ref{Eqn: ModelUpdates}). The pseudocode of the proposed FheFL is shown in Algorithms 1, 2 and 3. 


\begin{algorithm}[!ht]
\DontPrintSemicolon
  
  \KwInput{Users' Data and keys}
  \KwOutput{Global Model}
\tcc{Server Initiate the Process}
  {
  The server initialises an FL architecture \tcp*{}
  The server randomly initialises the weights of the architecture as $\mathbf{w}^0$ and sends this to all the users\tcp*{}
  }
\For{epoch $i$}
{
The server randomly selects a set of users\tcp*{}
The server let all the users know about the other users in the set  and send the global model $\mathbf{w}^i$\tcp*{}
    \For{each user $u$ in the set}
    {
    Compute the loss using the training data using (\ref{Eqn: Loss})\tcp*{}
    Calculate the gradients for each weight with respect to the loss and get $\nabla L_u(\mathbf{w}^i, \zeta^j_u)$\tcp*{}
    Encrypt $\nabla L_u(\mathbf{w}^i, \zeta^j_u)$ into $[\nabla L_u(\mathbf{w}^i, \zeta^j_u)]$ using the secret key $s_u$ and CKKS encryption algorithm \tcp*{}
    User $u$ sends $[\nabla L_u(\mathbf{w}^i, \zeta^j_u)]$ and shared secret key $ss_u = s_u + \sum_{j=1, j\neq u}^{U}s_{u,j}$ to the server  \tcp*{}
    }
The server performs secure aggregation using \textbf{Algorithm 3} to obtain $\mathbf{w}^{i+1}$ \tcp*{}
\If{ $||\mathbf{w}^{i+1}-\mathbf{w}^{i}|| \leq \epsilon $}
{
    Set Global model $= \mathbf{w}^{i+1}$ and \textbf{Exit} \tcp*{}
}
}
\caption{The proposed PPFL algorithm to mitigate Data Poisoning Attack}
\end{algorithm}

\begin{algorithm}[!ht]
\DontPrintSemicolon
  \KwInput{$[\nabla L_u(\mathbf{w}^i, \zeta^j_u)]$, $s_u+ \sum_{j=1, j\neq u}^{U}s_{u,j}$, and $\mathbf{w}^{i}$}
  \KwOutput{$\mathbf{w}^{i+1}$}
\tcc{Server performs the following}
\For{each user $u$ the Server}
{
Euclidean distance $[d^i_u]$ as in (\ref{eqn: ED Euclidean Distance }) \tcp*{$[d^i_u]$ is encrypted under user u's key $s_u$}
}
{
The server sum all $[d^i_u]$  to get $[\sum_{u=1}^{U}d^i_u]$ \tcp*{$[\sum_{u=1}^{U}d^i_u]$ is encrypted under the sum of all users' secret keys $\sum_{u=1}^{U}s_u$}
The server obtains $\sum_{u=1}^{U}s_u$ using  (\ref{Eqn: mk shared secret}) and decrypts $[\sum_{u=1}^{U}d^i_u]$ to get $\sum_{u=1}^{U}d^i_u$\tcp*{}

\For{each user $u$ the Server}
{
Computes non-poisoning rate $[p^i_u]$ using (\ref{Eqn: Poisoning rate in ED}) \tcp*{$[p^i_u]$ is encrypted by user u's secret key $s_u$}
Multiply $[p^i_u]$ and $[\nabla L_u(\mathbf{w}^i, \zeta^j_u)]$ to get $[p^i_u\cdot \nabla L_u(\mathbf{w}^i, \zeta^j_u)]$ \tcp*{$[p^i_u\cdot \nabla L_u(\mathbf{w}^i, \zeta^j_u)]$ is encrypted under user u's secret key $s_u$}
}
The server sum all $[p^i_u\cdot \nabla L_u(\mathbf{w}^i, \zeta^j_u)]$  to get $[\sum_{u=1}^{U}p^i_u\cdot \nabla L_u(\mathbf{w}^i, \zeta^j_u)]$ \tcp*{$[\sum_{u=1}^{U}p^i_u\cdot \nabla L_u(\mathbf{w}^i, \zeta^j_u)]$ is encrypted under the sum of all users' keys $\sum_{u=1}^{U}s_u$}
The server decrypts $[\sum_{u=1}^{U}p^i_u\cdot \nabla L_u(\mathbf{w}^i, \zeta^j_u)]$ using $\sum_{u=1}^{U}s_u$ to get $p^i_u\cdot \nabla L_u(\mathbf{w}^i, \zeta^j_u)$\tcp*{}
}
The server uses $\sum_{u=1}^{U}p^i_u\cdot \nabla L_u(\mathbf{w}^i, \zeta^j_u)$ and (\ref{Eqn: ModelUpdates}) to get $\mathbf{w}^{i+1}$.
\caption{The Proposed Secure Aggregation}
\end{algorithm}
 
 \section{Privacy, Security, and Convergence Analysis}\label{Section: Security and Privacy Analysis}
First, we establish a set of assumptions that form the foundation of the proposed FheFL scheme and subsequently conduct an analysis of privacy, security, and convergence aspects. We consider a scenario where certain users, referred to as colluding users, collude with both the server and other colluding users. These colluding users not only share their own secret keys with the server but also disclose the pairwise secret keys of other users. It is crucial to note that if there is only one non-colluding user, their model update can be compromised by the server. Therefore, our algorithm mandates the presence of at least two non-colluding users to ensure that only the sum of their respective model updates is accessible to the server. Increasing the number of non-colluding users enhances the level of protection against collusion. For the purposes of this study, we focus solely on data poisoning attacks, where malicious users deliberately retrain the model by introducing incorrect data samples that target specific classes. This type of attack is more detrimental than model poisoning attacks, as the intention is not to render the global model entirely useless, but rather to undermine its performance on the attacked classes while still maintaining reasonable performance on other classes which leads to a false sense of security. Finally, as discussed earlier, we assume a ratio between genuine and malicious users, whereby the percentage of malicious users is limited to $20\%$ \cite{FFL2023}. These assumptions provide a theoretical framework and context for the subsequent analysis of the proposed FheFL scheme.

 \subsection{Privacy Analysis}

\textbf{Corollary 1:} To compromise the data privacy of one benign non-colluding user, the aggregating server needs to collude with $(U-1)$ users.

\textit{Proof:} 
The server needs to compute the sum of $\sum_{u=1}^{U}s_u$, as shown in (\ref{Eqn: mk shared secret}), in order to decrypt the aggregated model. Assuming there are two non-colluding users, $u_1$ and $u_2$, the server has access to keys from the remaining $(U-2)$ users, except for $s_{u_1}$, $s_{u_2}$, and $s_{u_1,u_2}$. The server's objective is to decrypt the gradient updates, $[\nabla L_{u_1}(.)]$ and $[\nabla L_{u_2}(.)]$, provided by the two non-colluding users. It is important to note that these updates are encrypted using the CKKS FHE scheme with the respective users' keys, $s_{u_1}$ and $s_{u_2}$, and the server cannot directly decrypt them.

However, the server can obtain the aggregated model and subtract the model updates from colluding users. This operation only yields the sum of the model updates from the two non-colluding users. Similarly, the server computes the sum of the security keys, $s_{u_1} + s_{u_2}$. With this information, the server attempts to obtain the individual model updates, utilizing all available information, including the sum of the model updates from the non-colluding users. This allows the server to get the sum of the two non-colluding users' model updates. Hence, in order for the server to obtain the individual model updates from either of the two non-colluding servers, it is required to collude with one of them, necessitating collusion with $(U - 1)$ users.

Let us suppose that the server is successful in obtaining the individual model updates of the users then we can use this server to break the one-time pad encryption scheme which is based on information-theoretic security. We show a security game for this. Prior to that, let us define the information-theoretic one-time pad encryption.

\textit{Definition of information-theoretic one-time pad encryption:} Lets suppose there are two messages $m_0, m_1 \in \mathbb{Z}_q$. One-time pad encryption algorithm randomly selects a key $k \in \mathbb{Z}_q$ and encrypts the messages into ciphertext $c_0 = m_0+k \in \mathbb{Z}_q$ and   $c_1 = m_1+k \in \mathbb{Z}_q$. This encryption is information-theoretic secure because if you are given one of the ciphertexts then you won't be able to tell whether the ciphertext is encrypting $m_0$ or $m_1$ with probability more than $\frac{1}{2} + \epsilon$ where $\epsilon$ is very small. $\blacksquare$

\textbf{Security Game:} The aim of this security game is to use the aggregation server's ability to decrypt the models using the sum of the secret keys. We name this as server oracle, $\mathbb{SO}$ and define its function as follows: 
\begin{equation}
   \nonumber \mathbb{SO}([\nabla L_{u_1}(.)], s_{u_1}+s_{u_2})= \nabla L_{u_1}(.),
\end{equation}
where $[\nabla L_{u_1}(.)]$ is encrypted by $s_{u_1}$ and $s_{u_2}$ is generated independently of $s_{u_1}$. The output of $\mathbb{SO}$ is the decrypted $\nabla L_{u_1}(.)$.  

Similarly, let us also define an oracle, $\mathbb{OTP}$, that takes two messages and encrypts one of them using one-time padding as follows:
\begin{equation}
   \nonumber \mathbb{OTP}(m_1, m_2)= m_i+k \in \mathbb{Z}_q,
\end{equation}
where $i \in \{0,1\}$ and $k \in \mathbb{Z}_q$.  Now we can use $\mathbb{SO}$ and $\mathbb{OTP}$ to introduce the security game to break the information-theoretic of one-time pad encryption. To start the game, you select two messages $m_1$ and $m_2$ and pass them to $\mathbb{OTP}$. The output of $\mathbb{OTP}$ is $m_i+k \in \mathbb{Z}_q$.  Now select two messages, $\mathbf{x}^{1}$ and $\mathbf{x}^{1}$. Encrypt $\mathbf{x}^{1}$ using FHE using $m1$ as a secret key and get $[\mathbf{x}^{1}]$. Similarly, encrypt   $\mathbf{x}^{2}$ using FHE using $m2$ as a secret key and get $[\mathbf{x}^{2}]$. Now call  $\mathbb{SO}$ and input  $[\mathbf{x}^{1}]$, $[\mathbf{x}^{2}]$, and $m_i+k \in \mathbb{Z}_q$ as follows:
\begin{equation}\nonumber
    [\mathbf{x}^{1}, \mathbf{x}^{2}] \stackrel{?}{=} [\mathbb{SO}([\mathbf{x}^{1}], m_i+k), \mathbb{SO}([\mathbf{x}^{2}], m_i+k)].
\end{equation}
If $\mathbb{SO}$ correctly decrypts $[\mathbf{x}^{1}]$ then $m_i=m_1$ or if it is correctly decrypts  $[\mathbf{x}^{2}]$ then $m_i=m_2$.  Therefore, if the server can obtain the individual models then it can be used to break the information-theoretic security. For this reason, the proposed scheme protects the privacy of individual user models if there are two non-colluding users. $\blacksquare$

 \subsection{Security Analysis}

\textbf{Corollary 2:} The security of the proposed multi-key HE scheme is equivalent to the security of the FHE schemes.

\textit{Proof:} This proof is straightforward. If there is an adversary who can break the proposed multi-user HE scheme efficiently then the same adversary can be exploited to break the FHE efficiently. We can show this with a simple example. Let's suppose a message $m$ is encrypted using the FHE scheme as follows: $[m] = (a.s+m+e, a)$. To exploit the adversary, all we need to do is to homomorphically add $[m]$ several times (i.e., the number of users in the distributed network) and send it to the adversary. The adversary receives $\sum[m] = [m] + [m]+ \ldots = a(s+s+\ldots) + m + m + \ldots + e + e + \ldots$ which is very similar to the sum in the proposed multi-key HE algorithm. The adversary gets the sum of $m$, which can be divided to obtain $m$. Therefore, if an adversary can break the proposed multi-key homomorphic scheme easily compared to the FHE, then this server can be used to break the FHE schemes easily. For this reason, the security of the proposed scheme is the same as the security of FHE. $\blacksquare$

\subsection{Convergence Analysis}\label{Subsection: Convergence Analysis Proof}
In this section, we first show the intuition behind the proposed non-poisoning rate-based aggregation scheme. Then we derive an upper bound for the poisoning rate using theoretical convergence analysis. To facilitate this, let us denote the benign user set as $B$ and malicious users set as $M$. 

\textbf{Corollary 3:} If the sum of the non-poisoning rate for benign users is bigger than the malicious users i.e., $\sum_{b \in B} p_b^i > \sum_{m \in M}p_m^i$, then the proposed non-poisoning rate-based weighted aggregation converges to a model that is the same as the model obtained with benign users only.

\begin{figure}[!ht]
\centering
  \fbox{\includegraphics[height=5cm,clip]{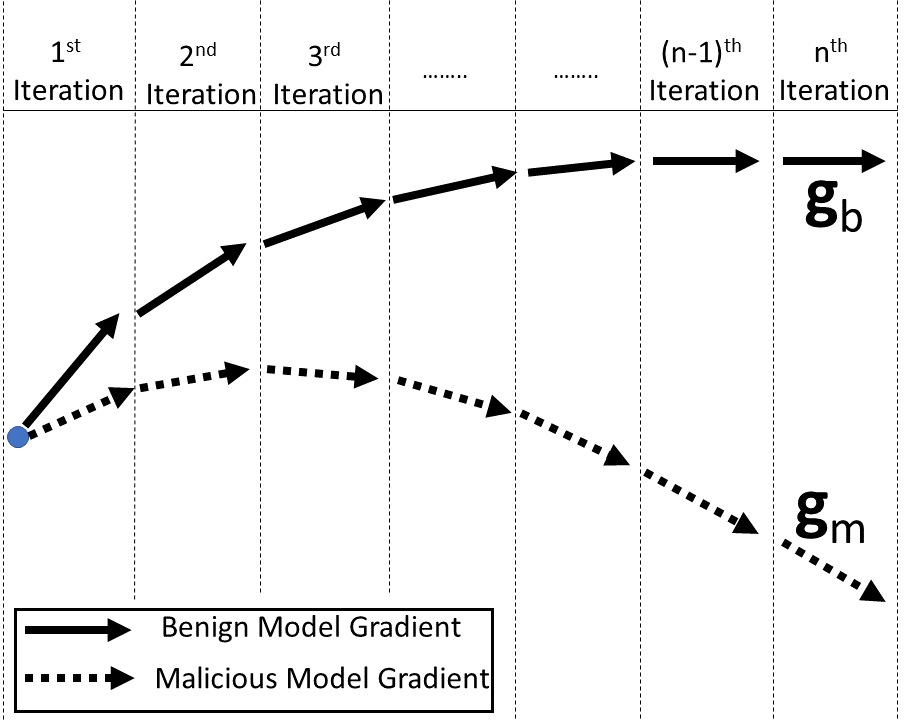}}
  \caption{Convergence of model updates when all the users are benign (solid arrows) or malicious (dotted arrows).}
  \label{Figure:Convergence}
\end{figure}

\textit{Intuition:} Let us consider two extreme scenarios.   The first scenario assumes all users are benign. Let \(\mathbf{g}_b^n = \frac{1}{|B|}\sum_{b \in B}\mathbf{w}_b^n\) be the global model obtained at the \textit{n}th epoch when all users are benign. The second scenario assumes that all the users are malicious, launching a label-flipping attack targeting a few classes. Let \(\mathbf{g}_m^n = \frac{1}{|M|}\sum_{m \in M}\mathbf{w}_m^n\) be the global model obtained at the \textit{n}th epoch when all users are malicious. Note that the FL algorithm converges in both scenarios \cite{ICLR2020}, but the directions of the final models obtained from these scenarios will be different. We can visualise this using a 2-dimensional example as in Figure \ref{Figure:Convergence}.

In a real setting, there will be both benign and malicious users.  Hence, the aggregated model after the \(i\)th iteration will be a model between \(\mathbf{g}_b^i\) and \(\mathbf{g}_m^i\). For brevity of notation, let's assume there is one malicious user and one benign user and the non-poisoning rate at the \(i\)th iteration for benign user as \(p_b^i\) and for malicious user as \(p_m^i\) (i.e., \(p_b^i + p_m^i = 1\)), then the aggregated model after the \(i\)th iteration can be represented as follows:
\[
\mathbf{g}^i =  p_b^i \mathbf{g}_b^i + p_m^i \mathbf{g}_m^i.
\]

If \(p_b^i > p_m^i\) then the aggregated model will be closer to \(\mathbf{g}_b^i\). The proposed non-poisoning rate-based weighted model aggregation minimizes the influence of malicious users in every iteration, i.e., \(p_b^{i+1} - p_m^{i+1} > p_b^i - p_m^i\). This ensures that as the iterations progress, the difference \(p_b^i - p_m^i\) becomes larger, leading to the increased weight of the benign model \(\mathbf{w}_b^i\) and the decreased weight of the malicious model \(\mathbf{w}_m^i\).

In the limit as \(i \to \infty\), we have \(p_b^i \to 1\) and \(p_m^i \to 0\). Thus,
\[
\lim_{i \to \infty} \mathbf{g}^i = \lim_{i \to \infty} p_b^i \mathbf{g}_b^i + p_m^i \mathbf{g}_m^i = \mathbf{g}_b^i.
\]

Therefore, as the number of iterations grows, the model \(\mathbf{g}\) converges to the benign users' model \(\mathbf{g}_b\), assuming that the non-poisoning rate-based weighted model aggregation is applied.

To formally prove Corollary 3, let us make the following assumptions (these assumptions are based on \cite{ICLR2020}) on the loss functions $L_1,\ldots, L_U$ (refer Section \ref{Section: Background}).

\textit{Assumption 1:} The loss functions $L_1,\ldots, L_U$ are all $L-$smooth if their gradients are Lipschitz continuous, i.e., for all $\mathbf{v}$ and $\mathbf{w}$:
\begin{equation}\label{Eqn: L-Smooth}
    \|\nabla L(\mathbf{v}) - \nabla L(\mathbf{w})\| \leq \alpha \|\mathbf{v} - \mathbf{w}\|.
\end{equation}
   
\textit{Assumption 2:} $L_1,\ldots, L_U$ are all $\mu-$strongly convex: for all $\mathbf{v}$ and $\mathbf{w}$:
\begin{equation}\label{Eqn: mu-strong}
    \nonumber L_u(\mathbf{v}) \geq L_u(\mathbf{w}) + (\mathbf{v}-\mathbf{w})^T \nabla L_u(\mathbf{w}) + \frac{\mu}{2} ||\mathbf{v}-\mathbf{w}||_2^2.
\end{equation}
\textbf{Proof:} To prove Corollary 3, we essentially need to show that if $p_b^i > p_m^i$ then $p_m^{i+1} < p_m^i$ i.e., the influence of malicious users would decrease with the iteration. As per (\ref{eqn: Euclidean Distance}) and (\ref{Eqn: Poisoning rate}), non-poisoning rate for (malicious) users $p_u^i$ inversely proportional to $||\nabla L_u(\mathbf{w}^i, \zeta^j_u)||_2^2$. For brevity, let's assume there are two users: 1) malicious user whose loss function is denoted as $L_m(.)$ and benign user whose loss function is denoted as $L_b(.)$. Also, assume that $p_m^i + p_b^i = 1$ where $p_m^i$ and $p_b^i$ are the non-poisoning rates for malicious and benign users at the $i$th iteration. Using these, we can get the following using (\ref{Eqn: ModelUpdates}):
\begin{equation}\label{Eqn: ModelUpdatesConverrgence}
    \mathbf{w}^{i+1} = \mathbf{w}^i - \eta^i \left[ p^i_m.\nabla L_m(\mathbf{w}^i) + p^i_b.\nabla L_b(\mathbf{w}^i)\right].
\end{equation}
Using (\ref{Eqn: ModelUpdatesConverrgence}) and (\ref{Eqn: L-Smooth}), we can obtain the following inequality for malicious user:
\begin{eqnarray}
   \nonumber  && \|\nabla L_m(\mathbf{w}^{i+1})  -  \nabla L_m(\mathbf{w}^{i})\| \leq \\
    && \alpha.\eta_i \|  p^i_m.\nabla L_m(\mathbf{w}^i) + p^i_b.\nabla L_b(\mathbf{w}^i) \|.
\end{eqnarray}
If $p^i_m < p^i_b$, then in the above inequality, the term on the right $\|  p^i_m.\nabla L_m(\mathbf{w}^i) + p^i_b.\nabla L_b(\mathbf{w}^i) \|$ is influenced by the benign user's gradient $\nabla L_b(\mathbf{w}^i)$ which may not be the same direction as the malicious user's gradient $\nabla L_m(\mathbf{w}^i)$. Therefore, if $p_b^i > p_m^i$ then $\|\nabla L_m(\mathbf{w}^{i+1}) - \nabla L_m(\mathbf{w}^{i})\| $ tend be bigger. Therefore, $\|\nabla L_m(\mathbf{w}^{i+1})\| > \| \nabla L_m(\mathbf{w}^{i})\| $.

We can observe the same phenomena using the $\mu-$strongly convex assumption. We can modify (\ref{Eqn: mu-strong}) for malicious user as follows:

\begin{eqnarray}
   \nonumber L_m(\mathbf{w}^{i+1}) \geq L_m(\mathbf{w}^i) &+& (\mathbf{w}^{i+1}-\mathbf{w}^i)^T \nabla L_m(\mathbf{w}^i)\\ \label{Eqn: mu-strongConver} &+& \frac{\mu}{2} ||\mathbf{w}^{i+1}-\mathbf{w}^i||_2^2.
\end{eqnarray}

For $(\mathbf{w}^{i+1}-\mathbf{w}^i)$ in (\ref{Eqn: mu-strongConver}), we can substitute $- \eta^i \left[ p^i_m.\nabla L_m(\mathbf{w}^i) + p^i_b.\nabla L_b(\mathbf{w}^i)\right]$ from (\ref{Eqn: ModelUpdatesConverrgence}). Therefore, if  \( p_m^i \) is small, the gradient term  (\(- \eta \nabla L_m(w^i)^\top \left( p^i_m.\nabla L_m(\mathbf{w}^i) + p^i_b.\nabla L_b(\mathbf{w}^i) \right)\)) in (\ref{Eqn: mu-strongConver}) influenced more by \( p^i_b.\nabla L_b(\mathbf{w}^i) \) than by \( p^i_m.\nabla L_m(\mathbf{w}^i)\). This can cause the model to move in a direction which is less aligned with minimizing \( L_m \), potentially increasing \( L_m(w^{i+1}) \). Similalrly, the quadratic term \(\left(\frac{\mu \eta^2}{2} \left\|  p^i_m.\nabla L_m(\mathbf{w}^i) + p^i_b.\nabla L_b(\mathbf{w}^i)  \right\|^2\right)\) in (\ref{Eqn: mu-strongConver}) penalizes large changes in the weights \( w^{i+1} - w^i \). Since \( p_m^i \) is small, \( p_b^i \) dominates, meaning the change in the loss \( L_m(w^{i+1}) \) might not align with the desired decrease from the malicious user's perspective. The penalty reinforces this by limiting how much the model can change in directions which is less favourable to the malicious user. Thus, under the assumptions of \(\alpha\)-smoothness and \(\mu\)-strong convexity, we can formally argue that \( p_m^{i+1} < p_m^i \) when \( p_m^i < p_b^i \). $\blacksquare$

Next, we need to derive the required conditions to achieve \( p_m^i < p_b^i \) so that the model converges in favour of benign users as proved in Corollary 3. To this, we make the following assumption:

\textit{Assumption 3:} The expected squared norm of stochastic gradients is uniformly bounded for benign users:
\begin{equation}\nonumber
E[||\nabla L_b(\mathbf{w}_t, \xi_t^b)||^2] \leq G^2, \forall b \in B.
\end{equation}
For data poisoning users, the expected squared norm of the poisoned gradients is bounded as follows:
\begin{equation}\nonumber
E[||\nabla L_m(\mathbf{w}_t, \xi_t^m)||^2] \leq G^2 + Z^2,  \forall m \in M,
\end{equation}
where $Z$ is a constant that bounds the impact of the poisoning attack on the expected squared norm of the gradient. The magnitude of $Z$ depends on the intensity of the data poisoning attack.

\textbf{Corollary 4:} If $Z^2 < \frac{(|B|-|M|)(|B|+|M|-1)}{|B||M|-|M|^2+|M|}.G^2.$ then $\sum_{m \in M} p_m < \sum_{b \in B} p_b$.

\textbf{Proof:} Using the Assumption 3, (\ref{eqn: Euclidean Distance}) and  (\ref{Eqn: Poisoning rate}), we can get the following inequalities:
\begin{eqnarray}
    \nonumber p_b &<& \frac{(|B|-1)G^2 + |M|(G^2+Z^2)}{|B|G^2 + |M|(G^2+Z^2)},\\
    \nonumber p_m &<& \frac{|B|G^2 + (|M|-1)(G^2+Z^2)}{|B|G^2 + |M|(G^2+Z^2)}.
\end{eqnarray}
We can obtain the required condition for $\sum_{m \in M} p_m < \sum_{b \in B} p_b$, by simplify $|M|p_m < |B|p_b$ as follows:
\begin{eqnarray}
    \nonumber |M|p_m &<& |B|p_b,\\
    \nonumber Z^2 &<& \frac{(|B|-|M|)(|B|+|M|-1)}{|B||M|-|M|^2+|M|}.G^2. \blacksquare
\end{eqnarray}

\section{Experimental Analysis}\label{Section: Experiments}
In this section, we report various experimental results to compare the performance of the proposed algorithm against the state-of-the-art (SOTA) algorithms. To conduct this experiment, we have implemented the proposed FheFL algorithm using the PyTorch framework and a modified version of the TenSEAL library \cite{Tenseal}. We used the MNIST, CIFAR-10 and CIFAR-100 datasets and The experiments were conducted on a 64-bit Windows PC with 16GB RAM and Intel(R) Core(TM) i7-4210U CPU at 4.1GHz.

\subsection{In Plain Domain}
First we experiment the proposed non-poisoning rate based aggregation scheme in plain domain and compare the results with SOTA plain domain algorithms such as [12][32][33].

\subsubsection{Dataset, neural network architecture, and users}
The MNIST, CIFAR-10, and CIFAR-100 datasets were used in our experiments. The MNIST dataset contains $70,000$ handwritten images of digits from $0$ to $9$, split into $60,000$ images for training and $10,000$ images for testing. Both CIFAR-10 and CIFAR-100 contain $60,000$ color images, with CIFAR-10 having 10 classes and CIFAR-100 having 100 classes. Each dataset is balanced with an equal number of images per class. In each case, the training dataset is evenly split and distributed across all the FL users. We considered a total of $100$ users, with 5\%, 15\%, and 20\% of them being malicious and conducting data poisoning attacks. For every epoch, the aggregation server randomly selects $10$ users for local model training, where a corresponding percentage of these selected users (either 1, 2, or 3 users) are malicious. For MNIST, a Convolutional Neural Network (CNN) was employed, featuring two convolution layers with 32 and 64 kernels of size 3 × 3, followed by a max-pooling layer and two fully connected layers with 9216 and 128 neurons. Meanwhile, CIFAR-10 and CIFAR-100 involved a more intricate model architecture, specifically the ResNet-18, chosen to accommodate the complexity of the dataset with its 10 classes and 100 class cases. This experimental design allowed us to comprehensively assess the models' capabilities across varied image classification challenges.

\subsubsection{Robustness Against Data Poisoning Attack}
In our experiments, we consider label-flipping attacks where malicious users poison data by changing the labels of particular classes and training the local model. The malicious participant intends to misclassify targeted labels during the prediction stage. The attackers were selected as participants with images of the digit "1" and then flipped the label from "1" to "7" for the MNIST dataset and vice versa. For CIFAR-10, the attackers were selected as participants while images of the "Cat" were then flipped to label from "Cat" to "Dog" and vice versa. For CIFAR-100, the attackers were selected as participants while images of the "Chair" were then flipped to label from "Chair" to "Couch" and vice versa. We conducted 200 synchronisation rounds for MNIST and CIFAR-10, and 500 synchronisation rounds for CIFAR-100. In each round of federated learning, participants were expected to train their local models for 5 epochs, while the attackers were allowed to train for an arbitrary number of epochs. In our experiments, we compared the proposed FheFL algorithm's accuracy against  FedAvg \cite{Fed_Avg_2017}, Trimmed Mean \cite{mean_median_2017}, Median \cite{mean_median_2017} and KRUM \cite{Krum2017}.  

\begin{table*}[!ht]
    \centering
    \caption{Accuracy comparison of the proposed FheFL's non-poisoning rate-based aggregation against the state-of-the-art schemes in the plain domain. We considered two scenarios for each dataset with 10 users and 100 users per round. }.
    \label{Table: Accuracy comparison}
    \begin{tabular}{|c|c|c|c|c|c|c|c|c|c|} 
    \multicolumn{7}{c}{ 0\% attackers} \\ \hline
    & \multicolumn{2}{c}{MNIST} \vline & \multicolumn{2}{c}{CIFAR-10} \vline & \multicolumn{2}{c}{CIFAR-100} \vline \\ \hline
        Algorithm                               & 10 & 100 & 10 & 100 & 10 & 100 \\ \hline
        FedAvg \cite{Fed_Avg_2017}              & \textbf{98.32\%} & 98.68\% & 97.57\% & 97.86\% & 78.13\% & 78.86\% \\ \hline
        Trimmed mean \cite{mean_median_2017}    & 94.57\% & 96.48\% & 93.04\% & 93.68\% & 77.54\% & 77.82\% \\ \hline
        Median \cite{mean_median_2017}          & 94.89\% & 96.88\% & 93.13\% & 93.92\% & 77.61\% & 77.90\% \\ \hline
        KRUM \cite{Krum2017}                    & 95.10\% & 96.34\% & 93.87\% & 94.14\% & 77.85\% & 78.08\% \\ \hline
        \textbf{FheFL (This work)}                      & 98.27\% & \textbf{98.78\%} & \textbf{97.86\% }& \textbf{98.07\%} & \textbf{78.28\%} & \textbf{78.92\%} \\ \hline
    \end{tabular}
\end{table*}
\begin{table*}[!ht]
    \centering
    \caption{Label Flipping attack stats. Average Accuracy and Average Attacker Success Rate(AASR) are illustrated against the popular aggregation algorithm.  This experiment was done only on the plain domain.}.
    \vspace{-4mm}
    \label{Table: Attack stat comparison}
    \begin{tabular}{|c|cc|cc|cc|}
        \multicolumn{7}{c}{ 5\% attackers} \\ \hline
          Dataset & \multicolumn{2}{c}{MNIST} \vline & \multicolumn{2}{c}{CIFAR-10} \vline & \multicolumn{2}{c}{CIFAR-100} \vline \\ \hline
        Algorithm                               & Accuracy & AASR &  Accuracy    & AASR & Accuracy   & AASR \\ \hline
        FedAvg \cite{Fed_Avg_2017}              & 97.55\%  & 0.55\% & 81.58\%  & 14.33\%  & 52.25\% & 7.34\% \\
        Trimmed mean \cite{mean_median_2017}    & 97.31\%  & 0.34\% & 79.16\%  & 13.63\%  & 51.20\% & 6.34\% \\
        Median \cite{mean_median_2017}          & 97.43\%  & 0.37\% & 80.01\%  & 13.23\%  & 52.03\% & 6.63\% \\
        KRUM \cite{Krum2017}                    & 96.63\%  & 0.42\% & 77.67\%  & 16.34\%  & 29.47\% & 8.34\% \\
        \textbf{FheFL (This work)}                       & 98.35\%  & 0.13\% & 84.87\%  & 2.31\%  & 59.35\% & 1.34\% \\ \hline
    
        \multicolumn{7}{c}{ } \\
        \multicolumn{7}{c}{ 15\% attackers} \\ \hline
        Dataset & \multicolumn{2}{c}{MNIST} \vline & \multicolumn{2}{c}{CIFAR-10} \vline & \multicolumn{2}{c}{CIFAR-100} \vline \\ \hline
        Algorithm                               & Accuracy & AASR &  Accuracy    & AASR & Accuracy   & AASR \\ \hline
        FedAvg \cite{Fed_Avg_2017}              & 92.56\%  & 2.05\% & 80.60\%  & 20.43\%  & 48.37\% & 58.65\% \\
        Trimmed mean \cite{mean_median_2017}    & 94.64\%  & 1.23\% & 81.16\%  & 18.34\%  & 49.17\% & 53.84\% \\
        Median \cite{mean_median_2017}          & 95.04\%  & 1.93\% & 82.53\%  & 17.4\%  & 50.30\% &  54.39\% \\
        KRUM \cite{Krum2017}                    & 93.09\%  & 3.63\% & 68.53\%  & 25.34\%  & 29.96\% & 64.71\% \\
        \textbf{FheFL (This work) }                      & 97.76\%  & 0.43\% & 83.65\%  & 3.44\%  & 55.45\% & 5.34\% \\ \hline

        \multicolumn{7}{c}{ } \\
        \multicolumn{7}{c}{ 20\% attackers} \\ \hline
        Dataset & \multicolumn{2}{c}{MNIST} \vline & \multicolumn{2}{c}{CIFAR-10} \vline & \multicolumn{2}{c}{CIFAR-100} \vline \\ \hline
        Algorithm                               & Accuracy & AASR &  Accuracy    & AASR & Accuracy   & AASR \\ \hline
        FedAvg \cite{Fed_Avg_2017}              & 84.54\% & 12.21\% & 80.15\%  & 22.58\%  & 46.78\% & 62.35\% \\
        Trimmed mean \cite{mean_median_2017}    & 87.21\% & 11.76\%  & 81.53\%  & 19.34\%  & 47.76\% & 61.79\% \\
        Median \cite{mean_median_2017}          & 88.72\% & 11.69\% & 82.48\%  & 19.03\%  & 47.65\% & 60.46\% \\
        KRUM \cite{Krum2017}                    & 86.93\% & 20.32\% & 69.05\%  & 26.69\%  & 28.65\% & 70.32\% \\
        \textbf{FheFL (This work)  }                     & 97.32\% & 1.78\% & 84.59\%  & 4.24\% & 52.85\% & 2.34\% \\ \hline
    \end{tabular}
\end{table*}

Table \ref{Table: Accuracy comparison} compares the accuracy of the FheFL against all other methods in plain domain without any attackers when 10 and 100 (all users) are participating in aggregation. The results show that when the number of users increases, model accuracy is generally going up. Moreover, the proposed non-poisoning rate-based weighted aggregation scheme outperforms Trimmed Mean \cite{mean_median_2017}, Mean Median \cite{mean_median_2017} and KRUM \cite{Krum2017} approaches. We also observed that our aggregation technique is slightly better than the FedAvg algorithm when the number of users is equivalent to $100$. This can be attributed to the non-poisoning rate which smoothes out the gradients which are far away from the intended direction.

\begin{figure*}[!ht]
  \centering
  \includegraphics[trim={0.8cm 0cm 0cm 0cm},width=0.95\textwidth]{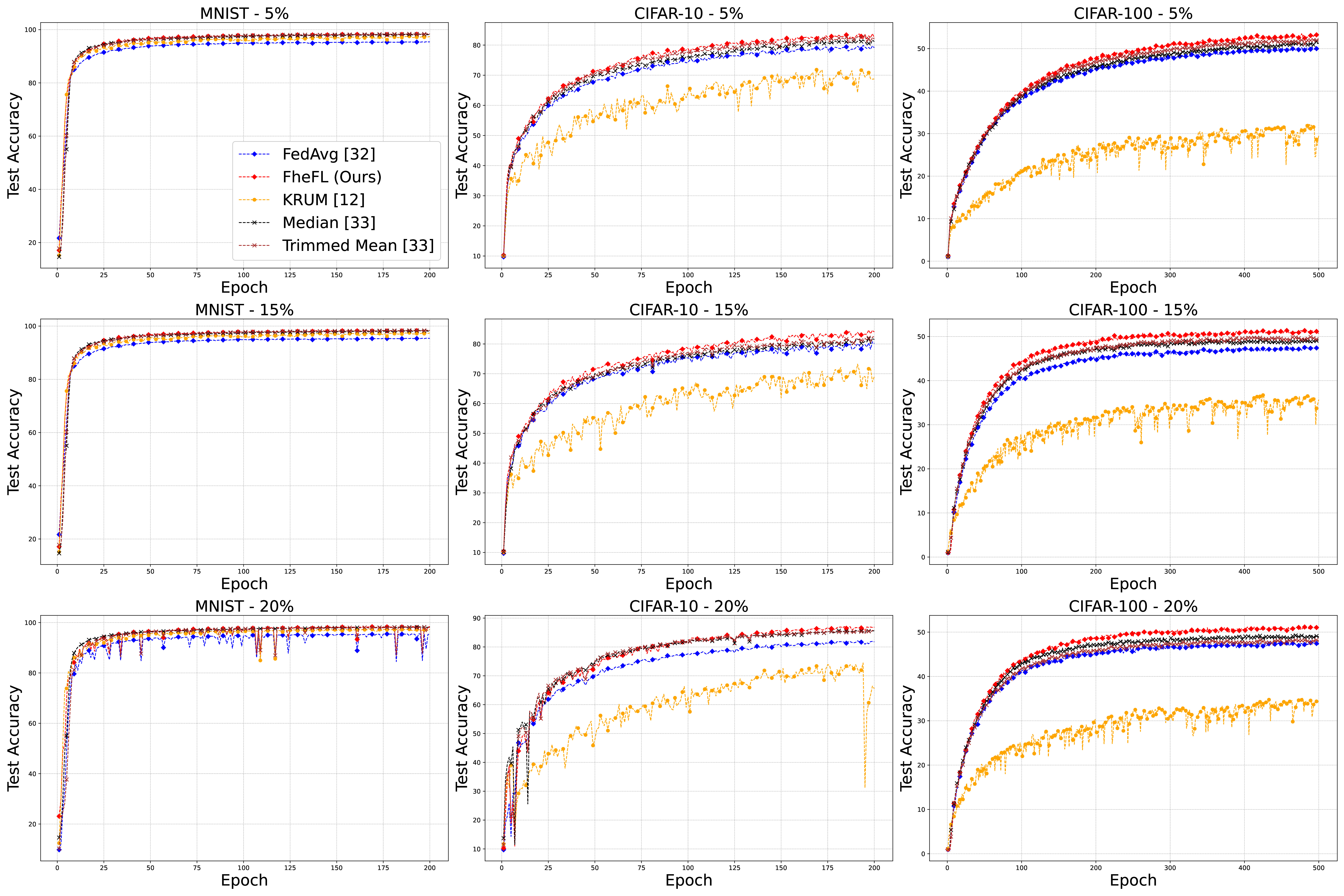}
  \vspace{4mm}
  \caption{This shows the accuracy of each aggregation algorithm when $5\%, 15\%$ and $20\%$ of the users are malicious for MNIST, CIFAR-10 and CIFAR-100 datasets}
  \label{Figure:label_flipping_10_acc}
\end{figure*}

\begin{figure*}[!ht]
  \centering
  \includegraphics[trim={0.8cm 0cm 0cm 0cm},width=0.9\textwidth]{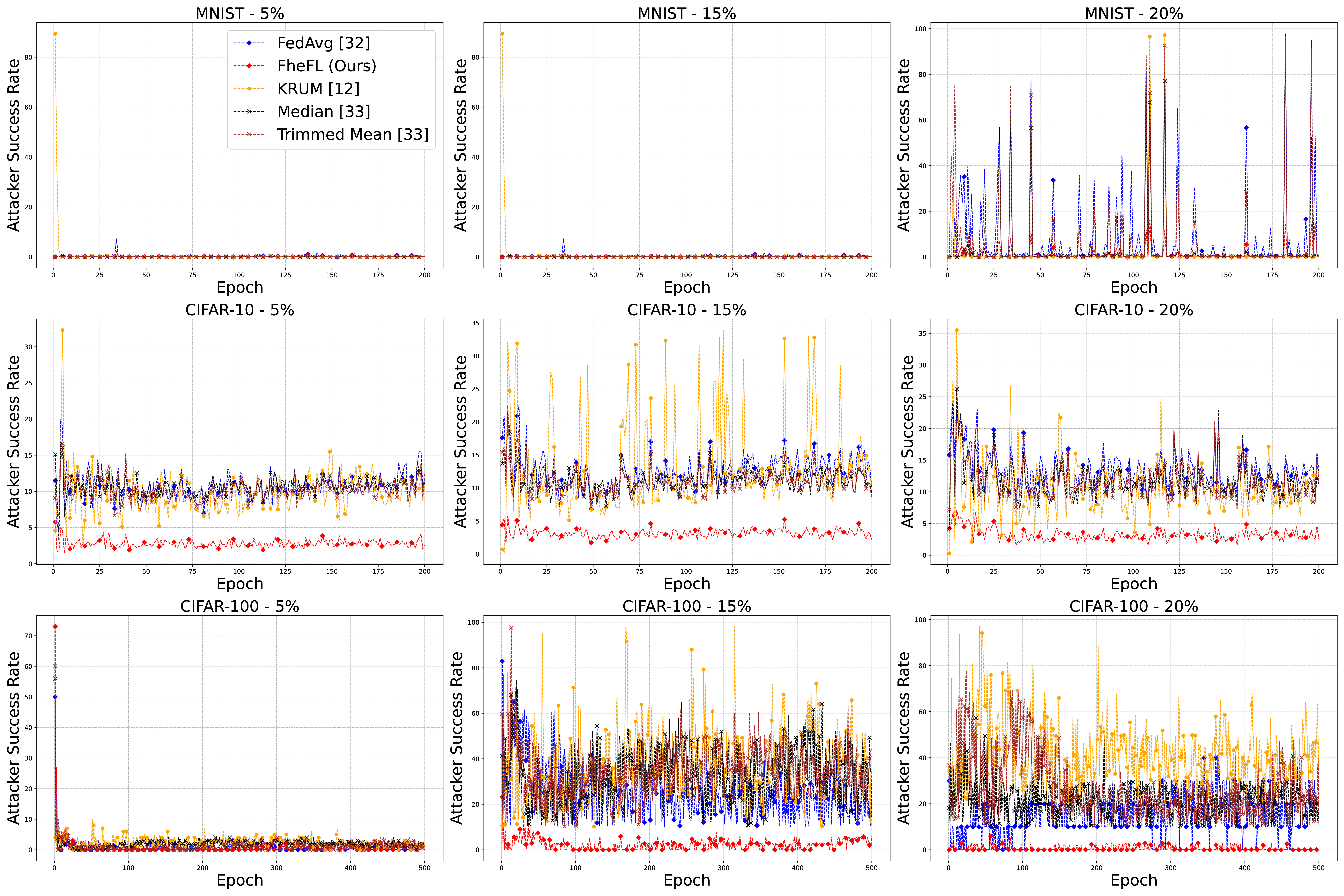}
   \vspace{2mm}
  \caption{Attacker Success rate for MNIST, CIFAR-10 and CIFAR-100 datasets, against, 5\%, 15\% and 20\% attackers}
  \label{Figure: Attacker success Rate}
\end{figure*}




To assess the robustness of our algorithms against malicious actors, we simulated a label-flipping attack with varying levels of adversarial participation. We gradually increased the proportion of malicious users from $0\%$ to $20\%$, specifically by introducing 5, 15, and finally 20 malicious users out of a total of 100. These malicious users were instructed to mislabel images based on the attack instructions explained previously. In each round of training, the server selected 10 users, ensuring that two of them were malicious.

Figure \ref{Figure:label_flipping_10_acc} illustrates the test accuracies achieved over successive training epochs for each of these scenarios across all three datasets (MNIST, CIFAR-10, and CIFAR-100).  Our proposed FheFL algorithm demonstrated comparable convergence rates to existing schemes. For instance, convergence was achieved around the 50th epoch for MNIST, between the 125th and 150th epochs for CIFAR-10, and between the 200th and 220th epochs for CIFAR-100. Notably, FheFL, with its non-poisoning rate-based aggregation scheme, consistently outperformed all other schemes in mitigating the impact of these label-flipping attacks.

The average accuracies under all three attack scenarios for all three datasets are provided in Table \ref{Table: Attack stat comparison}. On the same table, we show the Average Attack Success Rate (AASR). For example, for MNIST dataset, AASR is only $0.13\%$ for the proposed algorithm and notably, the proposed algorithm outperforms all the popular aggregation schemes in terms of overall accuracy and successfully mitigating the data poisoning attacks. When the percentage of attackers is increasing, the accuracy and detection rate are going down in all the algorithms which is expected. Moreover, we observed that FheFL has recorded a lower AASR in comparison to SOTA aggregation scheme during the data poisoning attacks. Furthermore, Figure \ref{Figure: Attacker success Rate} illustrates the AASR for all three attack scenarios for all three datasets.
\subsection{In Encrypted Domain}
Now we experiment the proposed algorithm in an encrypted domain and compare the computational and communications complexity against the other SOTA secure and privacy-preserving FL schemes such as [15][17[18][19].
\subsubsection{Key generation for CKKS FHE scheme}
The security key generation relies on several factors and depends on the underlying application. The high-level parameters $N$ and $q$ must be selected by considering efficiency and security. Moreover, the scaling factor $2^\Delta$ and a number of multiplication levels $L$ must be set in advance.  Since the application might be used by several users, the server presets these parameters common for all users. Given these global parameters ($N$, $q$, $\Delta$, and $L$), each user generates a public key and secret key pair. To achieve $128$-bit security, we considered two sets of parameters $(N=8192, q=218, \Delta=25,$ and $L=4)$ and $(N=16384, q=438,  \Delta=60,$ and $L=4)$. The proposed FL algorithm requires $L=4$ multiplication levels. When $N=16384$, we used $60-$bit precision, and for $N=8192$, the precision has been reduced to $25-$bit to accommodate $4-$levels of multiplication.

Additionally, with $N=8192$, users have the capacity to encrypt $4098$ gradients, and this number increases to $8192$ gradients when $N=16384$. Table \ref{Table: Model Complexity} provides a comprehensive overview of model complexity across all three datasets. As per Table \ref{Table: Model Complexity},  performing CKKS encryption on all weights necessitates transmitting numerous ciphertexts between the server and users, rendering this approach impractical.

However, previous research, notably Deep Leakage from Gradients
(DLG) \cite{DLG,iDLG}, indicates that encrypting or masking gradients can effectively hinder the server's ability to reconstruct data from shared gradients. Our experiments with DLG algorithms confirmed this and further revealed that encrypting gradients closer to the input layers provides superior mitigation against reconstruction compared to encrypting gradients closer to the output layers.

Figure \ref{Figure: DLG} shows some these observations when we attack the proposed FheFL algorithm using DLG on CIFAR100 dataset. The first row in Figure \ref{Figure: DLG} presents the original images from the CIFAR100 dataset. The second row displays reconstructed images without any gradient protection, highlighting the vulnerability to DLG attacks. The third row shows the reconstructions when the output layer's gradients are encrypted using our proposed scheme, while the final row presents the reconstructed images when the gradients of the first hidden layer are protected. The significant improvement in reconstruction quality when encrypting early-layer gradients underscores the effectiveness of this targeted approach. This is because gradients in the initial layers of a neural network capture low-level features and patterns in the training data, which are easier for attackers to reconstruct and exploit in DLG attacks. As shown in Table \ref{Table: Model Complexity}, the number of gradients in these early layers is significantly smaller than the overall complexity of the model. This allows for efficient encryption of these gradients using only one CKKS ciphertext, striking an ideal balance between security and computational overhead.

\begin{table}[h]
\centering
\caption{Comparison of the number of weights in the first convolutional layer and the total number of trainable weights for CNN architectures trained on MNIST, CIFAR-10, and CIFAR-100 datasets.}
\label{Table: Model Complexity}
\begin{tabular}{|c|c|c|}
\hline
Dataset & Weights in First Layer & Total Trainable Weights \\
\hline
MNIST & 320 & 11850 \\
\hline
CIFAR-10 & 2400 & 579402 \\
\hline
CIFAR-100 & 1728 & 9517378 \\
\hline
\end{tabular}
\end{table}

\begin{table}[h!]
\centering
\caption{The loss of accuracy between the plain domain and the FheFL with two sets of CKKS parameters,\cite{PEFL_2021} and \cite{ShieldFL_IEEE_TIFS_2022} .}
\label{Table: Accuracy LossX}
\begin{tabular}{|c|c|c|c|c|}
\hline
Number of Users & N=16384 & N=8192 & \cite{ShieldFL_IEEE_TIFS_2022}& \cite{PEFL_2021}\\ \hline
2  & 2.1  & 6.1 &4&3.5\\ \hline
50 & 2.3  & 6.2 &4&3.5\\ \hline
75 & 2.5  & 6.3 &4&3.5\\ \hline
100 & 2.7 & 6.5 &4&3.5\\ \hline
\end{tabular}
\end{table}

With a ring dimension of $N=8192$, we encrypt approximately $4000$ gradients closer to the input layers, and this increases to nearly $8000$ gradients when $N=16384$. It's important to acknowledge that CKKS encryption, which relies on approximate arithmetic, introduces a slight decrease in accuracy compared to plain domain computation. This accuracy loss is evident in Table \ref{Table: Accuracy LossX}, which compares the accuracy of our encrypted domain approach to the plain domain accuracy depicted in Figure \ref{Figure:label_flipping_10_acc}. Notably, with a larger ring dimension of $N=16384$ and a precision of $60$ bits, the accuracy loss remains below $3\%$ for one hundred users. However, reducing the ring dimension and precision to $N=8192$ and $25$ bits, respectively, results in doubling of the accuracy loss. While a smaller ring enhances efficiency, as detailed in the next subsection, it comes at the expense of accuracy. Similar trends have been observed in previous work  \cite{PEFL_2021} and \cite{ShieldFL_IEEE_TIFS_2022}. In  \cite{PEFL_2021}, where a plain neural network was employed, the accuracy loss between the plain domain and encrypted domain approaches was around $4\%$. Likewise, in \cite{ShieldFL_IEEE_TIFS_2022}, which utilized a simple logistic regression model, the accuracy difference between the plain domain and the encrypted domain was approximately $3.5\%$.

\subsection{Computational Complexity analysis}
\begin{figure}[h]
  \centering
  \includegraphics[width=0.4\textwidth]{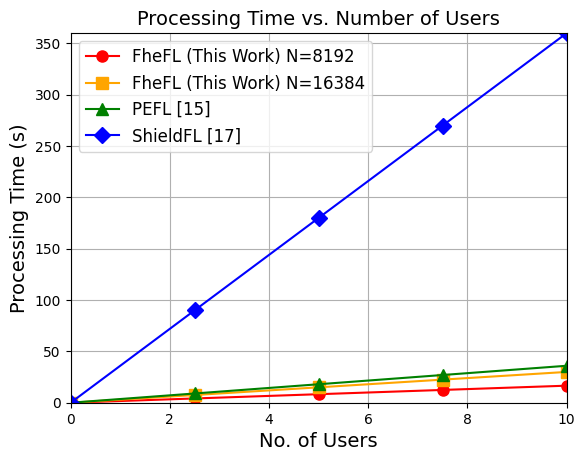}
   \vspace{-4mm}
  \caption{This shows the processing required by the server when the number of users is increasing. }
  \label{Figure: Processing Time  Vs Users}
\end{figure}

Figure \ref{Figure: Processing Time  Vs Users} comparing the computational time required by the server to perform the proposed aggregation in the encrypted domain against the SOTA schemes  \cite{PEFL_2021} and \cite{ShieldFL_IEEE_TIFS_2022}. It should be noted that the time complexity is measured based on the server that is sequentially processing the users' model updates. Calculating the Euclidean distance between the user's model update and the previous global model in the encrypted domain consumes a substantial amount of processing. However, this operation can be parallelized as it does not depend on the other users' input. Therefore, the time complexity shown in Figure \ref{Figure: Processing Time  Vs Users} can be decreased by the order of the users. Moreover, the computational time almost doubled when the server switched the CKKS security parameter $N$ from $8192$ to $16384$. While smaller $N$ is computationally efficient, this enforces a reduction in the number of bits preserved during the encrypted computation.


When $N=16384$, it took around 30 seconds for the proposed algorithm to aggregate model updates from 10 users where the number of elements in the model update is $8192$. However, as reported in \cite{ShieldFL_IEEE_TIFS_2022}, the ShieldFL algorithm requires more than one hour (the specification of the computer used in ShieldFL is very similar to the one we used for this paper as well as the number of gradients protected is same as the FheFL scheme). The main reason for this difference is that the proposed algorithm exploits the SIMD technique to pack $8192$ elements in one ciphertext. The PEFL algorithm in \cite{PEFL_2021} exploits a similar packing technique where it is expected to pack tens of elements in one plaintext, the time complexity can be reduced by around one order compared to the ShieldFL algorithm. Therefore, the proposed algorithm's complexity can be comparable to the algorithm in PEFL \cite{PEFL_2021}.

\subsection{Communicational Complexity Analysis}
\begin{figure}[!ht]
  \centering
  \includegraphics[trim={1cm 0cm 0cm 0cm}, width=0.4\textwidth]{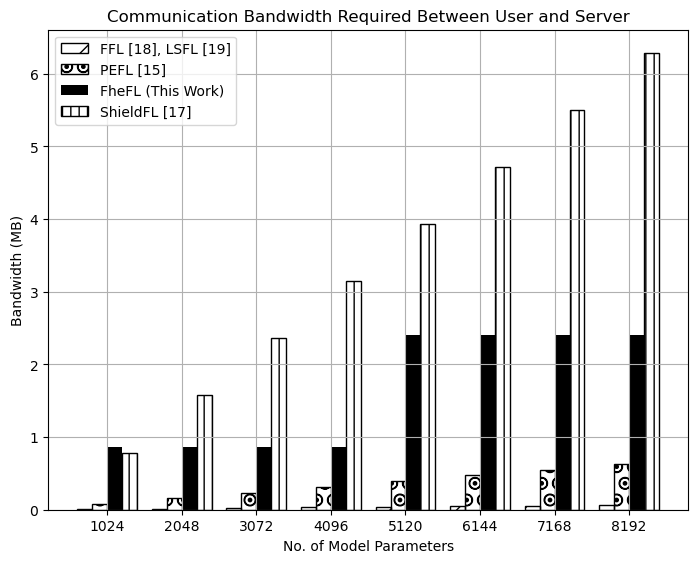}
  \vspace{-4mm}
  \caption{The bandwidth required for each user to transmit the model updates to the aggregation server.}
  \label{Figure: Bandwidth required}
\end{figure}

Figure \ref{Figure: Bandwidth required} compares the bandwidth requirement between a user and the server to communicate the updated model updates for different dimensions of the model updates. Since the schemes in \cite{FFL2023, LSFL} use one-time pad encryption, the bandwidth required to communicate the model updates is significantly lower compared to other schemes. The PEFL scheme in \cite{PEFL_2021} uses packing techniques to reduce the number of ciphertexts required. Therefore, the increase in communication bandwidth requirement for   \cite{PEFL_2021} compared to one-time pad encryption is not significant. Also, the bandwidth requirement is increasing linearly with the dimension of the model. The proposed algorithm uses the CKKS FHE scheme and its bandwidth requirement is almost an order higher than the \cite{PEFL_2021}. However, it should be noted that the proposed scheme relies on strong security assumptions and relies only on a single server. Nevertheless, each user requires less than $3MB$ bandwidth every epoch which is not significant given that the training can be conducted over Wi-Fi and/or 5G mobile network connectivity. The ShieldFl scheme in \cite{ShieldFL_IEEE_TIFS_2022} requires almost double the bandwidth compared to the proposed scheme. Moreover, \cite{ShieldFL_IEEE_TIFS_2022} requires two servers and the communication complexity between them is not accounted for in Figure \ref{Figure: Bandwidth required}.

\subsection{Convergence Analysis}\label{Subsection: Convergence Analysis Experiments}
\begin{figure}[!ht]
  \centering
  \includegraphics[trim={0.8cm 0cm 0cm 0cm}, width=0.4\textwidth]{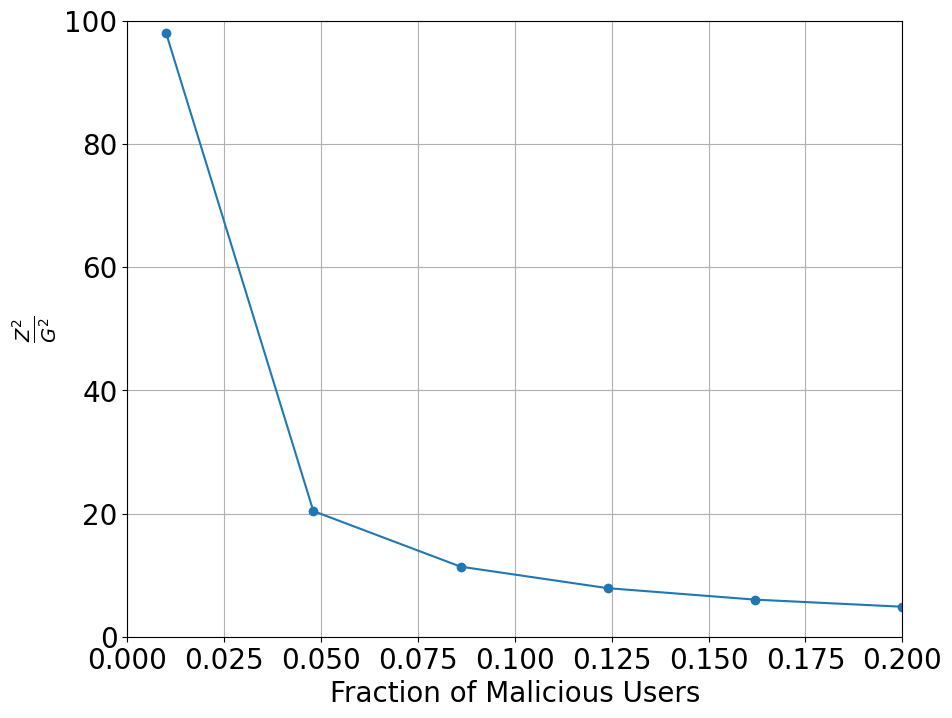}
  \caption{Upper bound for $Z^2$ to satisfy convergence.}
  \label{Figure: ZvsG}
\end{figure}

To confirm the convergence analysis proof in Subsection \ref{Subsection: Convergence Analysis Proof}, we measured training loss against epochs for varying numbers of malicious users. As established in Assumption 3 and Corollary 4, Figure \ref{Figure: ZvsG} illustrates the upper bound of $Z^2$ in relation to $G^2$ as the proportion of malicious users increases from $0\%$ to $20\%$. When $20\%$ of users are malicious, $Z^2$ can reach up to $10G^2$. Therefore, as per Assumption 3, for the malicious users, we have $E[||\nabla L_m(.)||^2] \leq G^2 + Z^2 = 11G^2$, whereas for benign users, $E[||\nabla L_b(.)||^2] \leq G^2$. This shows that the proposed FheFL framework can tolerate significantly higher variance—up to 11 times that of benign users. This level of variance exceeds what is typically observed in data poisoning attacks. Increasing the variance significantly might lead to \textit{model poisoning} instead of data poisoning. As explained earlier, data poisoning causes subtle changes in the accuracy of specific classes, leading to a false sense of security, as the model may appear effective during testing but underperforms or behaves maliciously when deployed in real-world applications.

\section{Conclusion}\label{Section: Conclusion}
In conclusion, the proposed FheFL scheme represents a significant advancement in federated learning by addressing privacy and security challenges. Leveraging FHE, FheFL ensures the privacy of user data by encrypting model updates and preventing privacy leakage. It introduces a distributed multi-key additive HE scheme for secure model aggregation, safeguarding individual user data during the model aggregation process. Additionally, a non-poisoning rate-based novel aggregation scheme combats data poisoning attacks in the encrypted domain. The scheme's security, privacy, convergence, and experimental analyses validate its effectiveness, offering comparable accuracy with reasonable computational costs while providing robust security guarantees. By introducing a novel framework with a single server and eliminating the need for user interaction in each epoch, FheFL sets the stage for future advancements in privacy-preserving federated learning.

\balance

\end{document}